\pdfoutput=1
\documentclass[11pt]{article}
\usepackage{acl}

\usepackage{enumitem}
\usepackage{booktabs,arydshln}
\usepackage[normalem]{ulem}
\usepackage{array}
\usepackage{hyperref}
\usepackage{url}
\usepackage{multirow}       
\usepackage{mdwlist}        
\usepackage{colortbl}
\usepackage{amssymb}
\usepackage{amsmath}
\usepackage{amsthm}
\usepackage{graphicx}
\usepackage{makecell}
\usepackage{threeparttable}
\usepackage{subcaption}
\usepackage{lscape}
\usepackage{fancyhdr}

\usepackage{wrapfig}
\usepackage{algpseudocode}
\usepackage{enumitem}
\usepackage{xcolor}	
\usepackage{algorithm}
\usepackage{mathtools}
\usepackage{times}
\usepackage{latexsym}
\usepackage{algorithmicx}
\usepackage{extarrows}
\usepackage{xspace}
\usepackage{longtable}
\usepackage{tikz}

\DeclareMathOperator*{\argmax}{arg\,max}

\makeatletter
\def\adl@drawiv#1#2#3{%
        \hskip.5\tabcolsep
        \xleaders#3{#2.5\@tempdimb #1{1}#2.5\@tempdimb}%
                #2\z@ plus1fil minus1fil\relax
        \hskip.5\tabcolsep}
\newcommand{\cdashlinelr}[1]{%
  \noalign{\vskip\aboverulesep
           \global\let\@dashdrawstore\adl@draw
           \global\let\adl@draw\adl@drawiv}
  \cdashline{#1}
  \noalign{\global\let\adl@draw\@dashdrawstore
           \vskip\belowrulesep}}
\makeatother

\newcolumntype{s}{>{\columncolor{blue!3}}r}
\newcolumntype{d}{>{\columncolor{yellow!4}}r}
\newcolumntype{q}{>{\columncolor{red!3}}r}

\usepackage{tabularx}
\usepackage{pgfplots}
\pgfplotsset{compat=newest}
\usepgfplotslibrary{groupplots,colorbrewer,statistics}
\usepackage{colortbl}
\definecolor{Azure}{RGB}{240,255,255}  
\definecolor{MistyRose}{RGB}{255,228,225} 
\definecolor{Ivory}{RGB}{255,255,240} 
\definecolor{Bisque}{RGB}{255,228,196}
\definecolor{Thistle}{RGB}{255,225,255}
\definecolor{orange}{RGB}{255,165,0}

\usepackage{url}

\renewcommand{\algorithmicrequire}{\textbf{Input:}}  
\renewcommand{\algorithmicensure}{\textbf{Output:}} 
\usepackage[T1]{fontenc}
\usepackage[utf8]{inputenc}
\usepackage{microtype}
\usepackage{inconsolata}
\title{
GumbelSoft: Diversified Language Model Watermarking via\\ the GumbelMax-trick
}

\author{
Jiayi Fu\textsuperscript{\rm $^\dagger$},
Xuandong Zhao\textsuperscript{\rm $^\ddagger$},
Ruihan Yang\textsuperscript{\rm $^\spadesuit$},\\
\bf Yuansen Zhang\textsuperscript{\rm $^\dagger$},
Jiangjie Chen\textsuperscript{\rm $^\dagger$},
Yanghua Xiao\textsuperscript{\rm $^\dagger$}\thanks{Corresponding author.}
\\
\textsuperscript{\rm $^\dagger$}Shanghai Key Laboratory of Data Science, School of Computer Science, Fudan University\\
\textsuperscript{\rm $^\spadesuit$}School of Data Science, Fudan University\\
\textsuperscript{\rm $^\ddagger$}University of California, Santa Barbara\\
\texttt{\{fujy22,rhyang21,zhangys22\}@m.fudan.edu.cn},
\\\texttt{xuandongzhao@ucsb.edu}, \texttt{\{jjchen19,shawyh\}@fudan.edu.cn}
}

\begin{document}
\maketitle

\begin{abstract}
Large language models (LLMs) excellently generate human-like text, but also raise concerns about misuse in fake news and academic dishonesty.
Decoding-based watermark, particularly the GumbelMax-trick-based watermark~(GM watermark), is a standout solution for safeguarding machine-generated texts due to its notable detectability. However, GM watermark encounters a major challenge with generation diversity, always yielding identical outputs for the same prompt, negatively impacting generation diversity and user experience. To overcome this limitation, we propose a new type of GM watermark, the Logits-Addition watermark, and its three variants, specifically designed to enhance diversity. Among these, the GumbelSoft watermark (a softmax variant of the Logits-Addition watermark) demonstrates superior performance in high diversity settings, with its AUROC score outperforming those of the two alternative variants by 0.1 to 0.3 and surpassing other decoding-based watermarking methods by a minimum of 0.1.\footnote{Code is available at \url{https://github.com/PorUna-byte/Gumbelsoft}}

\end{abstract}

\section{Introduction}
\label{section:intro}
The emergence of large language models (LLMs), exemplified by GPT-4 \citep{GPT4}, has enabled the generation of remarkably human-like content, facilitating tasks such as writing~\citep{shanahan2023evaluating}, coding~\citep{chen2021evaluating}, and fostering creativity. However, this technological advancement brings forth the potential for malicious applications, including social engineering~\citep{Mirsky_Demontis_Kotak_Shankar_Gelei_Yang_Zhang_Pintor_Lee_Elovici_etal._2023}, fake news fabrication~\citep{Zellers_Holtzman_Rashkin_Bisk_Farhadi_Roesner_Choi_2019}, and academic dishonesty. Consequently, the need for effective detection of machine-generated texts has become increasingly critical.

\begin{figure}[ht]
  \centering
  \includegraphics[width=1.0\linewidth]{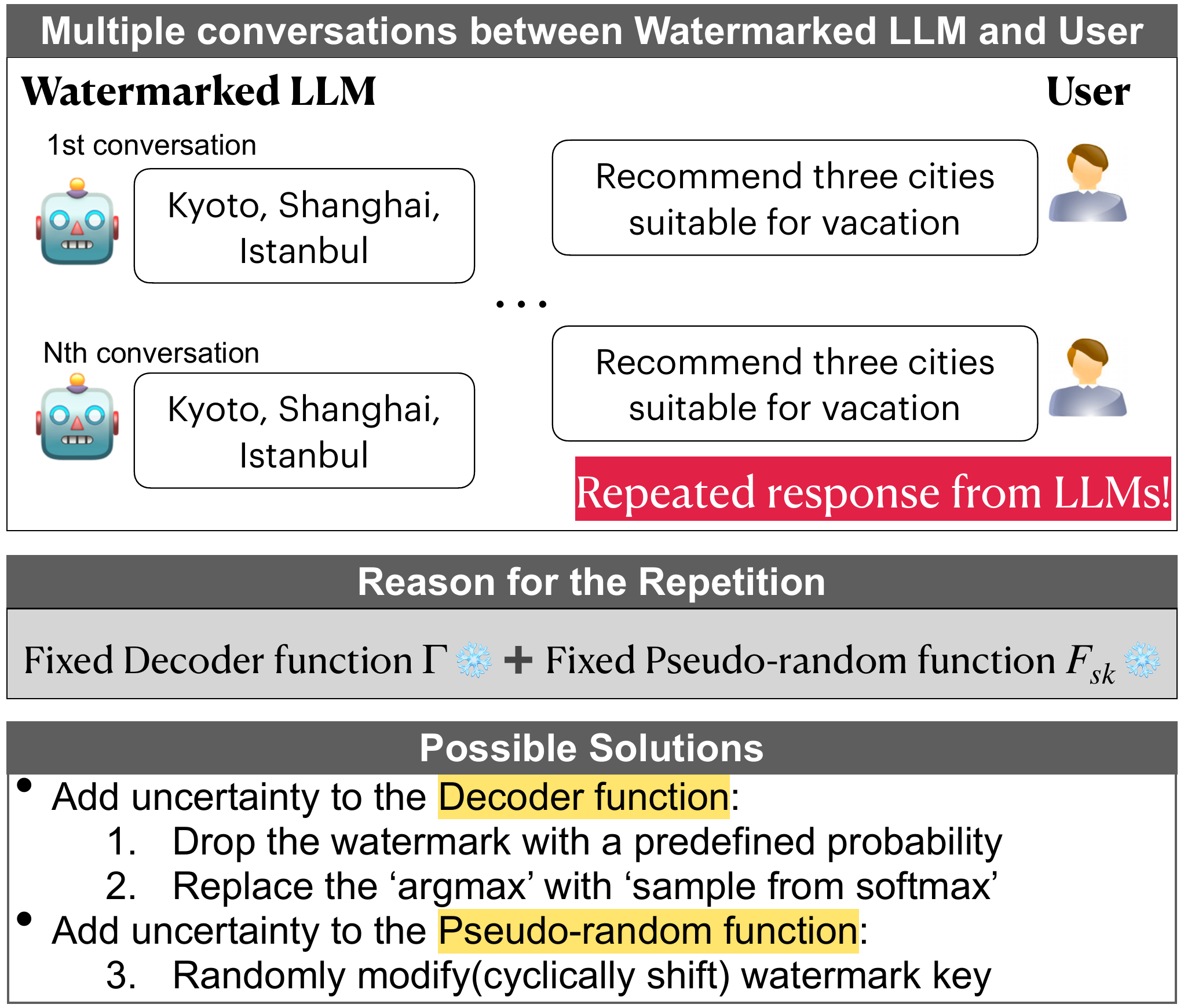}
  \caption{One significant limitation of GM watermark lies in their production of identical responses to the same queries. Such determinism can lead to user dissatisfaction, as individuals may become frustrated with LLM recommending the same outcomes for repeated prompts. This issue primarily stems from the deterministic nature of both the Pseudo-random function and the Decoder function. To address this concern, we propose three solutions: Solutions I and II aim to introduce variability into the Decoder function, whereas Solution III seeks to inject uncertainty into the Pseudo-random function.}
  \label{fig:problem_define_fig}
\end{figure}
Various strategies have been proposed to distinguish machine-generated texts from human-written texts, and decoding-based watermarking has emerged as a highly effective approach. 
This technique embeds subtle patterns into the text during the decoding stage of LLM,  which designated algorithms can identify.  
The GumbelMax-trick-based watermark (GM watermark), introduced by \citet{Aaronson_Kirchner} as their Exponential watermark, is a prominent example within this category, known for its exceptional detectability and low perplexity for generated text. 
However, a critical limitation of this method is its tendency to produce identical outputs for the same prompt, which could adversely affect both the diversity of the model's outputs and the overall user experience, as illustrated in Figure~\ref{fig:problem_define_fig}. 

To address the challenge of generating diverse outputs of the GM watermark, our analysis delves into the core mechanism of decoding-based watermarks. We discover that these watermarks share a cohesive framework, as illustrated in Figure~\ref{fig:decoding_fig}. The primary cause of uniform completions for identical prompts is traced back to the deterministic nature of both the \emph{Decoder} and \emph{Pseudo-random} functions in the GM watermark. To mitigate this, we propose two strategies to introduce variability into the Decoder function and one strategy to the Pseudo-random function:
1) Implement a drop mechanism with a predefined probability $d_p$, enabling direct sampling from the language model without watermark insertion.
2) Replace the ``argmax'' operation in GumbelMax watermark with ``sampling from softmax'' with temperature $\tau$.
3) Adjust the watermark key, derived from the Pseudo-random function, by cyclically shifting it $r$ positions—a method to effectively randomize the watermark key.

\emph{A critical aspect of this exploration is balancing detectability with diversity.} 
Integrating a dropout probability and shifting the watermark key boosts diversity but also reduces detectability. We propose replacing the argmax operation with ``sampling from softmax'' to enhance diversity without significantly compromising the watermark's integrity. This approach ensures that even though selections diverge from ``argmax'', they still achieve high per-token scores, preserving the statistical foundation of the watermark.
Further investigation into GM watermark leads us to question the necessity for an exponential transformation in the GumbelMax-trick for embedding watermarks, a technique outlined by~\citet{Aaronson_Kirchner}. Instead, we employ the GumbelMax-trick directly for watermark embedding and propose a distinct type of GM watermark, termed the Logits-Addition watermark.

Our experiments reveal that the GumbelSoft watermark, the softmax variant of the Logits-Addition watermark, consistently outperforms other GM watermark diversified variants in the AUROC metric, achieving a margin of 0.1 to 0.3 in high diversity settings. Additionally, the GumbelSoft watermark surpasses other decoding-based watermarks in AUROC by at least 0.1 on QA tasks, while maintaining low perplexity. 

\begin{figure}[t]
  \centering
  \includegraphics[width=1.0\linewidth]{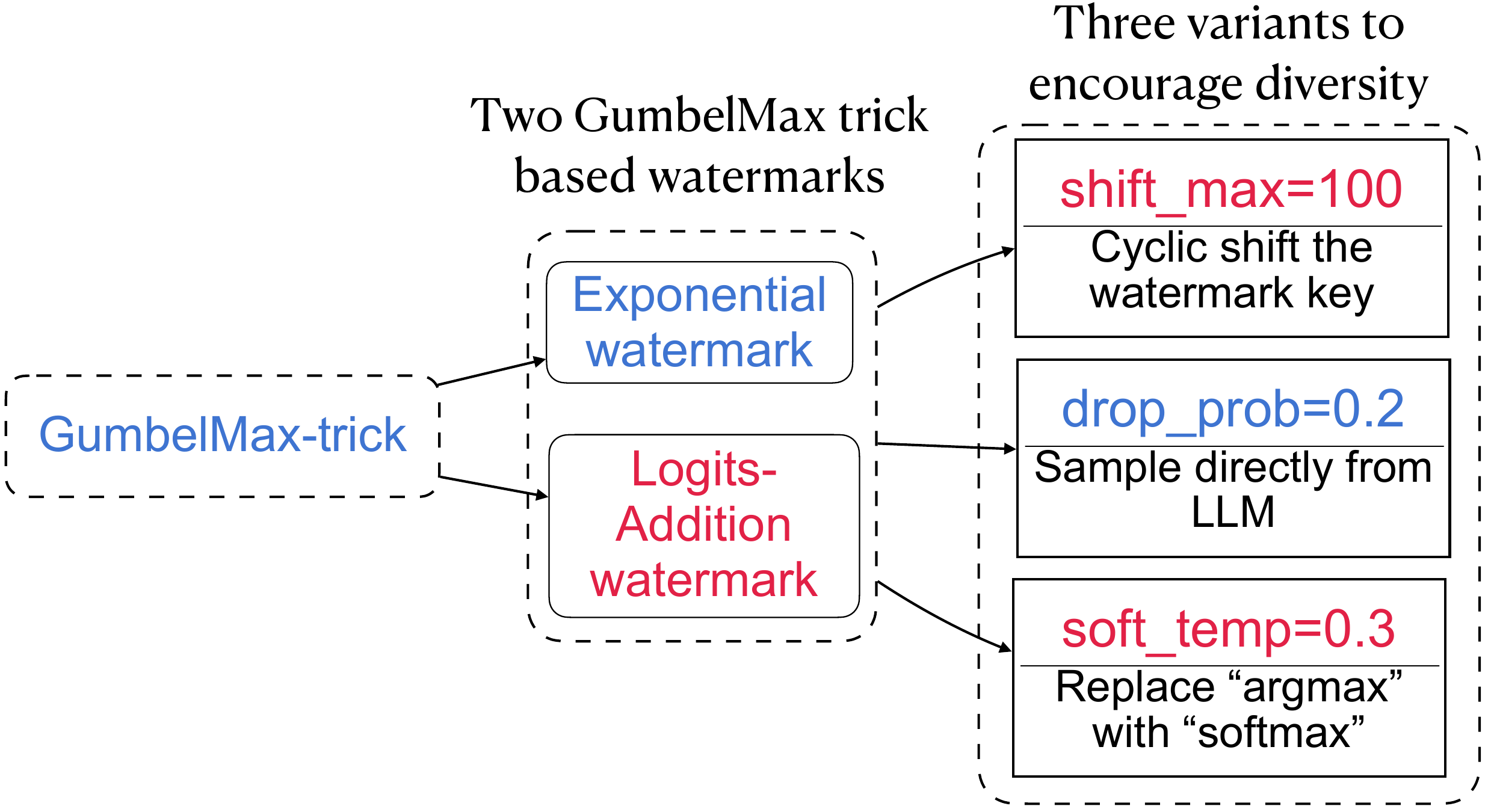}
  \caption{GumbelMax-trick can be used in text watermarking via two different ways: Exponential and Logits-Addition watermark. Each watermark has three variants to enhance generation diversity. The red part denotes our contribution, and the softmax variant of the Logits-Addition watermark is our suggested GumbelSoft watermark.}
  \label{fig:relationship}
\end{figure}

For a clearer understanding of these findings, we have illustrated the relationships among the GumbelMax-trick, the GM watermark (including Exponential and Logits-Addition), and their diversified variants in Figure~\ref{fig:relationship}.
In conclusion, our contributions are threefold:
\begin{itemize}[leftmargin=*]
\item We identify the deterministic nature of the Pseudo-random and Decoder functions as the primary cause behind GM watermark producing identical completions for the same prompts and provide a universal framework for all decoding-based watermarking techniques.
\item We propose the Logits-Addition watermark as a new type within the GM watermark suite and analyze the expectation and variance for the per-token score. Additionally, we introduce three variants of GM watermark aimed at enhancing the diversity of generated content.
\item Our experiments with three varied GM watermark versions reveal that the GumbelSoft watermark surpasses the others in diversity and detectability. Furthermore, our comparative analyses with other decoding-based watermarks show that the GM watermark offers superior detectability and robustness while maintaining quality on par with existing methods.
\end{itemize}

\section{Related Work}
\label{section:related}
Machine-generated text detection can be roughly categorized into three approaches.

\paragraph{Zero-shot Methods.} This approach, known as "model self-detection," necessitates full access to the language model and utilizes statistical measures such as perplexity and entropy. Noteworthy contributions include \citet{Gehrmann_Strobelt_Rush_2019}'s GLTR, \citet{vasilatos2023howkgpt}'s perplexity analysis, and \citet{yang2023dnagpt}'s N-gram overlaps. \citet{mitchell2023detectgpt} introduced a perturbation-based method, while \citet{deng2023efficient} proposed a Bayesian surrogate model. Recent studies encompass \citet{krishna2023paraphrasing}, who advocate for retrieval against paraphrase attacks, and \citet{su2023detectllm}, who leverage log-rank ratios. Additionally, \citet{solaiman2019release} employ log probability, \citet{bao2023fastdetectgpt} focus on conditional probability curvature, and \citet{venkatraman2023gptwho} utilize uniform information density for improved detection. The primary limitation of zero-shot methods lies in their requirement for complete access to the language model.

\paragraph{Training-Based Methods.} These involve classifiers trained to distinguish between machine-generated and human-written texts. \citet{chen2023gptsentinel,liu2023check} use a fine-tuned RoBERTa model \citep{Liu_Ott_Goyal_Du_Joshi_Chen_Levy_Lewis_Zettlemoyer_Stoyanov_2019}, while \citet{mireshghallah2023smaller} advocate for partially trained models. Some researchers also use shallow classifiers with extracted text features. \citet{openaidetect,gptzero} training classifiers from mixed sources, \citet{yin2023coco} using graph structures and contrastive learning, and \citet{tian2023multiscale} applying positive unlabeled training for classifier development.
\citep{li2023origin,tulchinskii2023intrinsic}. A drawback of training-based methods is their potential over-fitting to specific datasets and models.

\paragraph{Watermarking Techniques.} 
Recent advancements in hidden signal watermarking in texts can be categorized into post-edited and decoding-based watermarking. Post-edited watermarking involves text formatting or lexical changes \cite{Brassil_Low_Maxemchuk_O’Gorman_2002,sato2023embarrassingly,he2022cater,yoo2023robust}, while decoding-based watermarking in the era of large language models (LLMs) embeds statistical signals during decoding. Notable techniques in this domain include \citet{pmlr-v202-kirchenbauer23a}'s red-green list and \citet{Provable-robust}'s robust watermarking. Unbiased watermarks, which preserve the original token distributions, have been explored by \citet{kuditipudi2023robust} and \citet{hu2023unbiased}. Additionally, multi-bit watermarking, which embeds complex information, is examined by \citet{wang2023codable} and \citet{yoo2023advancing}. Several techniques for embedding watermarks include text formatting, as demonstrated by \citet{Por_Wong_Chee_2012} and \citet{Rizzo_Bertini_Montesi_2016}, and context-aware lexical substitution, as explored by \citet{yang2021tracing}. Syntactic modifications are discussed by \citet{Atallah2001_natural} and \citet{Meral_Sankur_Sumru}. Training data watermarking is addressed by \citet{liu2023watermarking} and \citet{tang2023did}. A publicly detectable watermark is proposed by \citet{Fairoze2023}, while leveraging semantic meaning for robustness is examined by \citet{ren2023robust}. \citet{zhao2024permute} propose PF-Watermark to further improve the text perplexity.
There are also some surveys on machine-generated content detection~\cite{wu2023survey, yang2023survey} and text watermarking~\cite{liu2024survey}.

\section{Method}
\label{section:method}
\begin{figure*}[ht]
  \centering
  \includegraphics[width=0.9\textwidth]{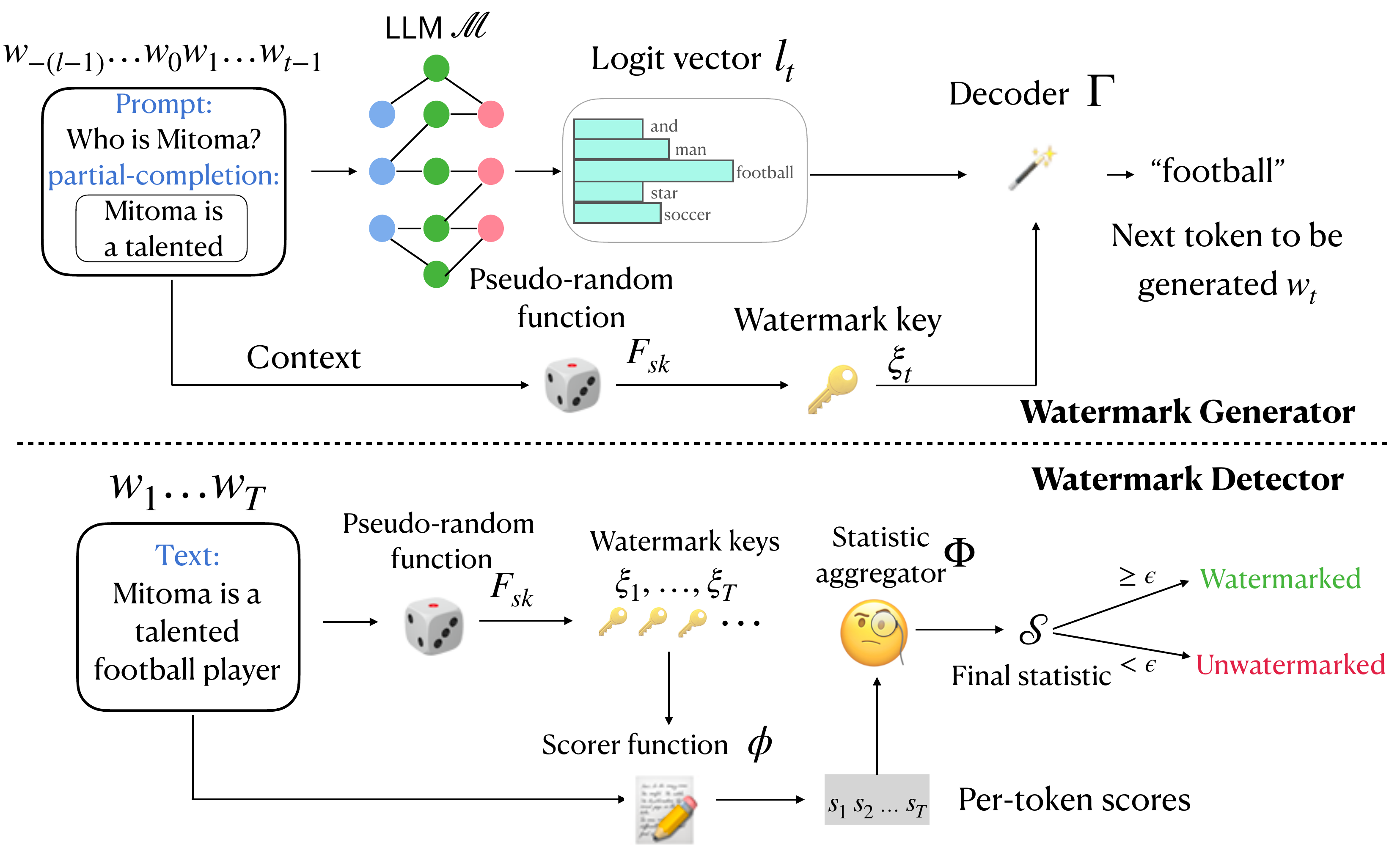}
  \caption{General framework of decoding-based watermark. The Generator uses logits vector $l_t$ and watermark key $\xi_t$ to decode the next token $w_t$. The Detector, employing scorer $\phi$, assesses the correlation between watermark key $\xi_t$ and token $w_t$, then combines these per-token scores to determine watermark presence. Both Generator and Detector share the same pseudo-random function $F_{sk}$. The context for watermark key calculation can be the preceding $h$ tokens.}
  \label{fig:decoding_fig}
\end{figure*}

\begin{table*}[ht!]
\centering
\small 
\setlength{\tabcolsep}{3.5pt}
\begin{tabular}{ll}
\toprule
\textbf{Symbol} & \textbf{Meaning} \\
\midrule
$\mathcal{V} $ & Vocabulary, the set of tokens \\
$w_t $ & Token at position t \\
$W_g:\mathcal{V}^*\rightarrow\mathcal{V}^*$ & Watermark Generator, generate a watermarked completion for a given prompt \\
$\mathcal{D}:\mathcal{V}^*\rightarrow\{\textrm{True}, \textrm{False}\}$ & Watermark Detector, detect whether a text is watermarked or not\\
$l_t\in\mathbb{R}^{|\mathcal{V}|}$ & Logits vector for position t, produced by language model $\mathcal{M}$\\
$\mathcal{M}:\mathcal{V}^*\rightarrow\mathbb{R}^{|\mathcal{V}|}$ & Language model, give the logits vector $l_t$ for position t based on a proceeding tokens\\
$\Xi$ & Watermark key space, the set of all possible watermark keys \\
$\xi_t\in\Xi$ & Watermark key at position t \\
$\mathcal{C}$ & Context space, the set of all possible contexts \\
$F_{sk}:\mathcal{C}\rightarrow\Xi$ & Pseudo-random function, calculate the watermark key $\xi_t$ \\
$\Gamma:\mathbb{R}^{|\mathcal{V}|}\times\Xi\rightarrow\mathcal{V}$ & Decoder function, decode the next token $w_t$ from logits vector and watermark key \\
$\phi:\mathcal{V}\times\Xi\rightarrow\mathbb{R}$ & Scorer function, calculate per-token score $s_t$ for each token\\
$\Phi:\mathbb{R}^*\rightarrow\mathbb{R}$ & Statistic aggregator, compile all per-token scores into one final statistic\\
\bottomrule
\end{tabular}
\caption{Summary of notations.}
\label{table:notation}
\end{table*}

In this section, we will first provide an overview of the decoding-based watermark framework and the GumbelMax-trick. Following this, we'll delve into the application of the GumbelMax-trick in text watermarking and examine their limitations. Concluding the section, we will present our recommended watermark scheme, specifically crafted to overcome these identified limitations.
\subsection{Preliminaries}
\paragraph{Decoding-Based Watermark Framework.}
We introduce a concise watermark framework with two main components: the Watermark Generator and Detector, building upon the architecture outlined in \citet{fernandez2023bricks} and incorporating mathematical concepts from \citet{kuditipudi2023robust,christ2023undetectable}. Figure~\ref{fig:decoding_fig} and Table~\ref{table:notation} detail the framework's structure and notations.

\paragraph{GumbelMax-trick.}
The GumbelMax-trick, as proposed by \citet{Gumbel}, presents an efficient method for sampling from a categorical distribution. Consider a vector of logits $l=(l_1,\ldots,l_K)$ coupled with a sequence of Gumbel-distributed random variables $g_1,\ldots, g_K \sim \textrm{Gumbel}(0,1)$. A sample from the categorical distribution $\pi=(\pi_1, \ldots, \pi_K)=\textrm{softmax}(l_1, \ldots, l_K)$ can be obtained as follows: $w=\argmax_{i}{(g_i+l_i)}$. This sampling approach is referred to as the GumbelMax-trick. It can be demonstrated that this trick is mathematically equivalent to drawing a sample directly from the categorical distribution $\pi$, as detailed in the Appendix~\ref{proof:Gumbel_unbiased}.

\subsection{Watermark Design}
\paragraph{Unbiasedness.}
The GumbelMax-trick enables the creation of an unbiased watermark, which is indistinguishable from unwatermarked text, provided the watermark key's distribution is properly chosen. An unbiased watermark meets the following conditions:
$$\mathbb{P}_{\xi\sim \tau(\cdot)}[\Gamma(\xi,l)=x]=p_x, \forall x\in\mathcal{V}$$
where $p$ is the softmax of $l$ and $\tau(.)$ denotes the watermark key $\xi$'s distribution. Watermark schemes in \citet{Aaronson_Kirchner,wu2023dipmark,kuditipudi2023robust} are unbiased, unlike the biased method of \citet{pmlr-v202-kirchenbauer23a}. 
\paragraph{Logits-Addition Watermark.}
The first attempt to use GumbelMax-trick in text watermarking is \citet{Aaronson_Kirchner}'s Exponential watermark, which generates subsequent tokens using the formula $w_t=\argmax_{i}\frac{\log\xi_{t}[i]}{p_{t}[i]}$, where $\xi_t\sim \textrm{Uniform}(0,1)^{|\mathcal{V}|}$ and $p_t=\textrm{softmax}(l_t)$. Its detection mechanism computes a per-token score $s_t=-\log(1-\xi_t[w_t])$. 

While the Exponential watermark is linked to empirical entropy, we question the relevance of this connection given that empirical entropy does not accurately reflect the true entropy of the next-token distribution provided by the language model. Consequently, we introduce a new type of GM watermark that directly incorporates Gumbel noise into the logits vector for next-token sampling: $w_t = \argmax_{i} {(l_t[i] + \xi_t[i])}$, where $\xi_t \sim \text{Gumbel}(0,1)^{|\mathcal{V}|}$ and $l_t$ represents the logit vector. This method's detection algorithm calculates a per-token score $s_t = \xi_t[w_t]$, a technique we designate as the Logits-Addition Watermark.

We assert that despite the token generation processes of these two methods being equivalent (see Appendix~\ref{proof:Gumbel_equivalent}), their detection mechanisms differ. Furthermore, the softmax variant of our Logits-Addition watermark demonstrates superior diversity and detectability compared to the Exponential watermark's softmax variant (refer to Figure~\ref{fig:diversity_auroc}). This supports our rationale for applying Gumbel noise directly and adopting an alternative detection method. Moreover, we present a theorem detailing the expectation and variance of the per-token score within the Logits-Addition watermark.

\newtheorem{theorem}{Theorem}
\begin{theorem}
Consider a text $w_1, \ldots, w_T$ embedded with a watermark using the Logits-Addition technique. When evaluated by the Logits-Addition watermark detector, the expected value and variance of the score for each token are given by
\begin{equation*}
    \begin{aligned}
    \mathbb{E} [s_t] = & \mathbb{E} [\xi_t[w_t]] = -\log(p_t[w_t])+\gamma,\\
    \textrm{Var}\left[s_t\right] \leq& \frac{2 p_t\left[w_t\right]^2}{\left(1-p_t\left[w_t\right]\right)^3}+\frac{2}{p_t\left[w_t\right]}\\
    &-\left(-\log p_t\left[w_t\right]+\gamma\right)^2.
    \end{aligned}
\end{equation*}
For a non-watermarked text $w_1, \ldots, w_T$, applying the Logits-Addition watermark detector, the expected value and variance for each per-token score are
\begin{equation*}
    \begin{aligned}
        &\mathbb{E} [s_t] = \mathbb{E} [\xi_t[w_t]] = \gamma,\\
        &\textrm{Var}[s_t]= \textrm{Var}[\xi_t[w_t]] = \frac{\pi^2}{6}.
    \end{aligned}
\end{equation*}
Here, $\gamma$ denotes the Euler-Mascheroni constant, and $p_t = \text{softmax}(l_t)$, is derived from the language model.
\label{theorem:expectation of GumbelMax watermark}
\end{theorem}
The proof for this theorem can be found in Appendix~\ref{proof:Gumbel_expectation}.
According to this theorem, if certain watermarked tokens are assigned a low probability by the language model, the expectation of their per-token scores, given by $-\log(p_t[w_t]) + \gamma$, significantly increases. This makes these tokens notably easier to detect.

\paragraph{Limitations of the GM Watermark.}
Despite its effectiveness in watermarking texts, the GumbelMax-trick has limitations. One major limitation is that it generates deterministic outputs, resulting in identical completions for the same prompts (as shown in Figure~\ref{fig:problem_define_fig}). Such determinism can lead to user dissatisfaction, as individuals may become frustrated with LLM consistently recommending the same outcomes for the same queries. To address this issue and improve output diversity, we propose three diversified GM watermark variants. These variants are thoroughly outlined in the Introduction section (see Section~\ref{section:intro}) and are aimed at enhancing the diversity of the generation process.

\subsection{GumbelSoft Watermark}
\begin{algorithm}[ht]
\caption{GumbelSoft Generator}
\label{algorithm_1}
\begin{algorithmic}[1]
\Require prompt $x$, LLM $\mathcal{M}$, temperature $\tau$.
\Ensure 
    Watermarked completion $w_1, \ldots, w_T$
\For{$t = 1, \ldots, T$}
    \State Logits \(l_t\leftarrow \mathcal{M}(x,w_{1,\ldots,t-1})\)
    \State Watermark key \(\xi_t\leftarrow\)hash context to a Gumbel-distributed vector
    \State \(w_t\leftarrow\)sample from $\textrm{softmax}((\xi_t+l_t)/\tau)$
\EndFor\\
\Return $[w_1, \ldots, w_T]$ 
\end{algorithmic}
\end{algorithm}

\begin{algorithm}[ht]
\caption{GumbelSoft Detector}
\label{algorithm_2}
\begin{algorithmic}[1]
\Require 
     Text input $w_{1,\ldots,T}$;
     a predefined threshold $\epsilon$
\Ensure 
     Boolean indicator: True if watermark detected, False otherwise
\For{$t = 1, \ldots, T$}
    \State Watermark key \(\xi_t\leftarrow\)hash context to a Gumbel-distributed vector
    \State Per-token score \(s_t\leftarrow\xi_t[w_t]\)
\EndFor
\State Calculate Final statistic $\mathcal{S}$: $$\mathcal{S}=\Phi(s_1,s_2,\ldots,s_T)=\frac{\sqrt{6T}}{\pi}(\frac{\sum_{i=1}^Ts_i}{T}-\gamma)$$ with $\gamma\approx 0.5772$ denoting the Euler-Mascheroni constant. \\
\Return True if $\mathcal{S}\geq\epsilon$ else False. 
\end{algorithmic}
\end{algorithm}
After conducting a comprehensive series of experiments with three diversified variants of both the Exponential and Logits-Addition watermarks, we identified the GumbelSoft watermark as the most effective, achieving Pareto optimality. The methodologies for both the Generator and Detector of the GumbelSoft watermark are elaborated in Algorithms~\ref{algorithm_1} and \ref{algorithm_2}, respectively.

We now explain the key insight behind the GumbelSoft watermark. The Logits-Addition watermark is primarily characterized by differing expected per-token scores for watermarked and unwatermarked texts. Leveraging this difference allows for the construction of a detection mechanism based on the null hypothesis $\mathcal{H}_0$: \textit{The text is unwatermarked}. 
Following the z-test by~\citet{pmlr-v202-kirchenbauer23a}, we devise the final statistic $\mathcal{S}$ of Logits-Addition watermark to be:$$\mathcal{S}=\Phi(s_1,s_2,\ldots,s_T)=\frac{\sqrt{6T}}{\pi}\left(\frac{\sum_{i=1}^Ts_i}{T}-\gamma\right)$$
According to expectation Theorem~\ref{theorem:expectation of GumbelMax watermark} and the central limit 
Theorem~\citep{central_limit_theorem}, we notice that for unwatermarked texts, $\mathcal{S}$ aligns with a standard Gaussian distribution. In contrast, for watermarked texts, $\mathcal{S}$ deviates, typically presenting significantly higher values. 
Given that GumbelSoft is a variant of Logits-Addition, it naturally inherits its characteristics. Consequently, the majority of tokens sampled by the GumbelSoft watermark are likely identical to those selected by the Logits-Addition watermark. Moreover, tokens not usually favored by Logits-Addition are observed to have comparatively higher per-token scores.

\section{Experiment}
\label{section:experiment}
This section presents a comparative study of three diversified variants (refer to Figure~\ref{fig:problem_define_fig}) of both Exponential and Logits-Addition watermarks, with an emphasis on aspects such as detectability, diversity, and quality. Following this, the optimal diversified variant of the GM watermark, the GumbelSoft watermark, is identified and compared against several existing decoding-based watermark schemes (see Appendix~\ref{proof:baselines} for details).
\subsection{Experimental Setting}
We briefly outline our experimental setup, including the datasets, models, metrics, and baselines used, specifically for the Completion and QA tasks.
\paragraph{Dataset and Models.}
In our experimental setup, each task employs unique language models and datasets. For the Completion task, the Llama2-7b model~\citep{touvron2023llama} and C4 dataset~\citep{Raffel_Shazeer_Roberts_Lee_Narang_Matena_Zhou_Li_Liu_2019} are used to assess detectability, while diversity is evaluated through 20 high-entropy prompts repeated 50 times each on Llama2-7b. Perplexity is calculated using Llama2-13b with the C4 dataset. For the QA task, we utilize the Llama2-7b-chat model and Alpaca dataset~\citep{alpaca} for detectability, and assess diversity with 20 chat-like prompts on Llama2-7b-chat, also repeated 50 times. Perplexity here is measured using Llama2-13b-chat on the Alpaca dataset.
\begin{figure}[t]
\centering
\small
\begin{tikzpicture}
  \begin{groupplot}[
    group style={
      group size=2 by 1,
      ylabels at=edge left,
      yticklabels at=edge left,
    },
    width=6cm,
    height=5cm,
    scatter/classes={
        a1={mark=triangle*, color of colormap={200}, mark size=4pt},
        a2={mark=triangle*, color of colormap={400}, mark size=4pt},
        a3={mark=triangle*, color of colormap={600}, mark size=4pt},
        a4={mark=triangle*, color of colormap={800}, mark size=4pt},
        a5={mark=triangle*, color of colormap={1000}, mark size=4pt},
        b1={mark=diamond*, color of colormap={200}, mark size=4pt},
        b2={mark=diamond*, color of colormap={400}, mark size=4pt},
        b3={mark=diamond*, color of colormap={600}, mark size=4pt},
        b4={mark=diamond*, color of colormap={800}, mark size=4pt},
        b5={mark=diamond*, color of colormap={1000}, mark size=4pt}
    },
    legend style={
      at={(1,0.2)},
      anchor=east,
      legend columns=1
    },
  ]
  \nextgroupplot[
    xlabel=Self-Bleu (better $\rightarrow$),
    ylabel=AUROC (better $\rightarrow$),
    colorbar,
    colormap/viridis,
    colorbar style={
        at={(1.05,1)},               
        anchor=north west,        height=\pgfkeysvalueof{/pgfplots/parent axis height},
        title={$\tau$},
        point meta min=0, 
        point meta max=0.5, 
        ytick={0.1,0.2,0.3,0.4,0.5},
    },
    xtick={0.65,0.7,0.75,0.8,0.85,0.9},
    x dir=reverse,
    xmin=0.65, xmax=0.9,
    ymin=0.88, ymax=0.92,
    ytick={0.88,0.89,0.90,0.91,0.92},
    ymajorgrids=true,
    xmajorgrids=true,
  ]
  \addlegendimage{only marks, mark=triangle*, gray} 
  \addlegendimage{only marks, mark=diamond*, gray} 
  \addplot[scatter,only marks,
            scatter src=explicit symbolic] 
    table[meta=label] { 
    x       y       label
    0.89    0.907   a1
    0.803   0.91    a2
    0.745   0.911   a3
    0.713   0.914   a4
    0.68    0.912   a5
    0.9     0.907   b1
    0.811   0.904   b2
    0.782   0.901   b3
    0.755   0.9     b4
    0.736   0.898   b5
    };
    \addlegendentry{GumbelSoft}
    \addlegendentry{Exp+softmax}
\end{groupplot}
\end{tikzpicture}
\caption{The figure shows how AUROC changes with Self-Bleu on the QA task. we use different colors to represent temperature and different marks to represent GumbelSoft and the softmax variant of Exponential watermarks. The AUROC is calculated for 100 detection tokens. Since the top-right outshines the bottom-left in performance, GumbelSoft is more effective than the softmax variant of Exponential.}
\label{fig:diversity_auroc}
\end{figure}
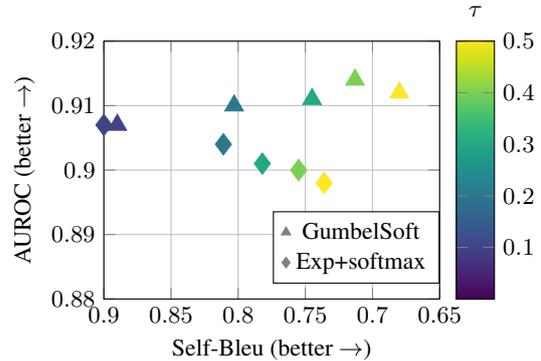
\paragraph{Metrics.}
Our detectability evaluation relies on AUROC, FPR at a fixed FNR of 0.01, and FNR at a fixed FPR of 0.01. We assess generation quality using perplexity, derived from a larger model. To measure generation diversity, our approach includes Self-BLEU and Distinct 1-gram and 2-gram.
\paragraph{Baselines.}
The universal decoding-based watermark framework, as presented in Figure~\ref{fig:decoding_fig}, serves to categorize all decoding-based watermark schemes, including those proposed by~\citet{pmlr-v202-kirchenbauer23a,Aaronson_Kirchner,wu2023dipmark,kuditipudi2023robust}. These schemes are the baselines in our study. Their mathematical representations, provided in Appendix~\ref{proof:baselines}, illustrate their integration into our unified taxonomy.
\subsection{Diversity}
\begin{table*}[t]
\centering
\small
\begin{tabular}{ccsssdddq}
\toprule
& & \multicolumn{3}{c}{\textbf{Diversity}}  & \multicolumn{3}{c}{\textbf{Detectability}} & \multicolumn{1}{c}{\textbf{Quality}} \\
\cmidrule(lr){3-5} \cmidrule(lr){6-8} \cmidrule(lr){9-9}
& & \multicolumn{1}{c}{\textbf{Self-Bleu} $\downarrow$}  & \multicolumn{1}{c}{\cellcolor{white}\textbf{Dist-1} $\uparrow$}&   \multicolumn{1}{c}{\textbf{Dist-2} $\uparrow$} &  \multicolumn{1}{c}{\textbf{AUROC} $\uparrow$} &  \multicolumn{1}{c}{\textbf{FPR} $\downarrow$ }&  \multicolumn{1}{c}{\textbf{FNR} $\downarrow$ }&  \multicolumn{1}{c}{\textbf{PPL} $\downarrow$ } \\

\midrule
\multirow{15}{*}{\rotatebox{90}{\textbf{Exponential}}}
&vanilla & 1.000 & 0.011 & 0.017 & 0.905 & 0.749 & 0.569 & 1.985 \\
\cdashlinelr{2-9}
&drop\_prob=0.10 & 0.852 & 0.070 & 0.196 & 0.891 & 0.790 & 0.623 & 2.020 \\
&drop\_prob=0.20 & 0.767 & 0.087 & 0.261 & 0.871 & 0.835 & 0.691 & 2.015 \\
&drop\_prob=0.30 & 0.715 & 0.097 & 0.298 & 0.845 & 0.870 & 0.752 & 2.077 \\
&drop\_prob=0.40 & \textbf{0.676} & 0.103 & 0.325 & 0.816 & 0.896 & 0.808 & 2.000 \\
\cdashlinelr{2-9}
&shift\_max=30 & 0.902 & 0.080 & 0.227 & 0.742 & 0.946 & 0.825 & 1.996 \\
&shift\_max=50 & 0.839 & 0.090 & 0.266 & 0.700 & 0.963 & 0.882 & 1.985 \\
&shift\_max=100 & 0.741 & 0.101 & 0.311 & 0.672 & 0.963 & 0.900 & 1.982 \\
&shift\_max=200 & 0.689 & \textbf{0.106} & \textbf{0.331} & 0.644 & 0.970 & 0.901 & 1.983 \\
\cdashlinelr{2-9}
&soft\_temp=0.2 & 0.811 & 0.084 & 0.233 & 0.904 & 0.748 & 0.586 & 2.372 \\
&soft\_temp=0.3 & 0.782 & 0.087 & 0.254 & 0.901 & 0.756 & 0.597 & 2.096 \\
&soft\_temp=0.4 & 0.755 & 0.094 & 0.276 & 0.900& 0.794 & 0.598 & 2.239 \\
&soft\_temp=0.5 & 0.736 & 0.096 & 0.288 & 0.898 & 0.798 & 0.602 & 2.127 \\
\midrule
\multirow{15}{*}{\rotatebox{90}{\textbf{Logits-Addition}}}
&vanilla & 1.000 & 0.011 & 0.017 & 0.908 & 0.743 & 0.579 & 1.985 \\
\cdashlinelr{2-9}
&drop\_prob=0.10 & 0.823 & 0.074 & 0.212 & 0.887 & 0.769 & 0.634 & 1.998 \\
&drop\_prob=0.20 & 0.762 & 0.089 & 0.263 & 0.867 & 0.830 & 0.701 & 1.994 \\
&drop\_prob=0.30 & 0.713 & 0.097 & 0.300 & 0.846 & 0.833 & 0.748 & 2.088 \\
&drop\_prob=0.40 & 0.691 & 0.102 & 0.316 & 0.810 & 0.888 & 0.808 & 1.988 \\
\cdashlinelr{2-9}
&shift\_max=30 & 0.906 & 0.080 & 0.224 & 0.730 & 0.961 & 0.838 & 1.986 \\
&shift\_max=50 & 0.824 & 0.092 & 0.272 & 0.694 & 0.965 & 0.886 & 1.986 \\
&shift\_max=100 & 0.751 & 0.101 & 0.309 & 0.670 & 0.971 & 0.903 & 1.981 \\
&shift\_max=200 & 0.694 & \textbf{0.106} & \textbf{0.331} & 0.642 & 0.981 & 0.917 & 1.981 \\
\cdashlinelr{2-9}
&soft\_temp=0.2 & 0.803 & 0.083 & 0.235 & 0.910 & 0.726 & \textbf{0.568} & 2.338 \\
&soft\_temp=0.3 & 0.745 & 0.095 & 0.281 & 0.911 & \textbf{0.704} & 0.572 & 2.027 \\
&soft\_temp=0.4 & 0.713 & 0.098 & 0.300 & \textbf{0.914} & 0.713 & 0.570 & 2.169 \\
&soft\_temp=0.5 & 0.680 & 0.105 & 0.326 & 0.912 & 0.742 & 0.571 & 2.221 \\
\bottomrule
\end{tabular}
\caption{Comparison of three diversified variants of both Exponential and Logits-Addition watermarks in the QA task. These variants include \textbf{drop\_prob=0.2}, sampling from the language model directly at a 0.2 probability; \textbf{shift\_max=100}, where the watermark key is cyclically shifted within a 0-100 range; and \textbf{soft\_temp=0.3}, which uses a softmax sampling with a temperature of 0.3 to balance randomness.
\textbf{Vanilla} is the original GM watermark(Exponential and Logits-Addition) without any technique to enhance diversity. The detectability is measured by 100 detection tokens. Note that Logits-Addition+soft\_temp is the GumbelSoft watermark. GumbelSoft is the best of three diversified variants of GM watermark in terms of both detectability and diversity.}
\label{table:diversity_chat}
\end{table*}
This subsection aims to identify which variant of the two GM watermarks is best in terms of diversity and detectability.
A detailed comparison of our GumbelSoft watermark with other GM watermark variants in the QA task is presented in Table~\ref{table:diversity_chat}, with results for the Completion task detailed in Appendix~\ref{Appendix:diversity}. These results indicate that our GumbelSoft method achieves superior content diversity and detectability compared to other variants, though it incurs a slight increase in perplexity. We also notice that the GumbelSoft watermark is better than the softmax variant of the Exponential watermark under the same temperature setting, which is clearly shown in Figure~\ref{fig:diversity_auroc}.
\begin{table*}[t]
\centering
\resizebox{\textwidth}{!}{ 
\small
\begin{tabular}{ccsssdddqqqq}
\toprule
& & \multicolumn{3}{c}{\textbf{\# tokens=40}}  & \multicolumn{3}{c}{\textbf{\# tokens=60}} & \multicolumn{4}{c}{\textbf{\# tokens=100}} \\
\cmidrule(lr){3-5} \cmidrule(lr){6-8} \cmidrule(lr){9-12}& 
& \multicolumn{1}{c}{\textbf{AUROC} $\uparrow$}  
& \multicolumn{1}{c}{\cellcolor{white}\textbf{FPR} $\downarrow$} 
& \multicolumn{1}{c}{\textbf{FNR} $\downarrow$} 
& \multicolumn{1}{c}{\textbf{AUROC} $\uparrow$}  
& \multicolumn{1}{c}{\cellcolor{white}\textbf{FPR} $\downarrow$}
& \multicolumn{1}{c}{\textbf{FNR} $\downarrow$} 
& \multicolumn{1}{c}{\textbf{AUROC} $\uparrow$} 
& \multicolumn{1}{c}{\textbf{FPR} $\downarrow$ }
& \multicolumn{1}{c}{\textbf{FNR} $\downarrow$ }
& \multicolumn{1}{c}{\textbf{PPL} $\downarrow$ } \\
 
\midrule
\multirow{6}{*}{\rotatebox{90}{\textbf{Completion}}}
& Unwatermarked & - & - & - & - & - & - & - & - & - & 11.576 \\
& KGW & 0.970 & 0.616 & 0.361 & 0.988 & 0.329 & 0.164 & 0.997 & 0.078 & 0.041 & 14.217\\
& Exponential & 0.997 & 0.012 & 0.012 & 0.999 & \textbf{0.000} & 0.006 & \textbf{1.000} & \textbf{0.000} & \textbf{0.000} & 10.953\\
& Dipmark & 0.935 & 0.693 & 0.565 & 0.968 & 0.483 & 0.362 & 0.988 & 0.274 & 0.153 & \textbf{8.664} \\
& ITS & 0.961 & 0.073 & 1.000 & 0.978 & 0.040 & 1.000 & 0.994 & 0.010 & 0.402 & 11.843 \\
& \textbf{GumbelSoft} & \textbf{0.998} & \textbf{0.011} & \textbf{0.010} & \textbf{1.000} & \textbf{0.000} & \textbf{0.005} & \textbf{1.000} & \textbf{0.000} & 0.001 & 11.820 \\

\midrule
\multirow{6}{*}{\rotatebox{90}{\textbf{QA}}}
& Unwatermarked & - & - & - & - & - & - & - & - & - & 1.980 \\
& KGW & 0.657 & 0.985 & 0.969 & 0.701 & 0.978 & 0.945 & 0.754 & 0.949 & 0.901 & 2.081\\
& Exponential & 0.780 & 0.892 & 0.813 & 0.840 & 0.852 & 0.738 & 0.905 & 0.749 & \textbf{0.569} & 1.985 \\
& Dipmark & 0.588 & 0.988 & 0.982 & 0.615 & 0.981 & 0.984 & 0.646 & 0.979 & 0.970 & \textbf{1.792} \\
& ITS & 0.583 & 1.000 & 1.000 & 0.618 & 0.963 & 1.000 & 0.665 & 0.954 & 1.000 & 2.011 \\
& \textbf{GumbelSoft} & \textbf{0.788} & \textbf{0.866} & \textbf{0.812} & \textbf{0.848} & \textbf{0.837} & \textbf{0.722} & \textbf{0.911} & \textbf{0.704} & 0.572 & 2.027 \\

\bottomrule
\end{tabular}
}
\caption{A comparative analysis of the detectability across various decoding-based watermarking schemes. Detectability is assessed for varying token counts: 40, 60, and 100. The temperature for GumbelSoft is set to 0.3. GumbelSoft shows high detectability and low perplexity compared with other decoding-based watermarks.}
\label{table:detectability}
\end{table*}

While methods like drop probability and watermark key shift can enhance diversity, they tend to negatively impact detectability. The decrease in detectability due to drop probability may be attributed to a fraction of tokens not being sampled using the watermark key, thereby diluting the overall statistical strength. In the case of shifted watermark keys, the detection phase becomes more complex as every possible shift must be tested to identify the watermark, potentially leading to inflated statistics for unwatermarked texts and thus reducing detectability. In contrast, our GumbelSoft watermark does not encounter these issues, maintaining high detectability while also enhancing generation diversity.
\subsection{Detectability and Quality}
This subsection aims to show that the GumbelSoft watermark is better than other decoding-based watermarks in terms of detectability, the results are shown in Table~\ref{table:detectability} and the hyperparameter is detailed in Appendix~\ref{Appendix:detectability}. 
\begin{figure}[t]
\centering
\includegraphics[width=1.0\linewidth]{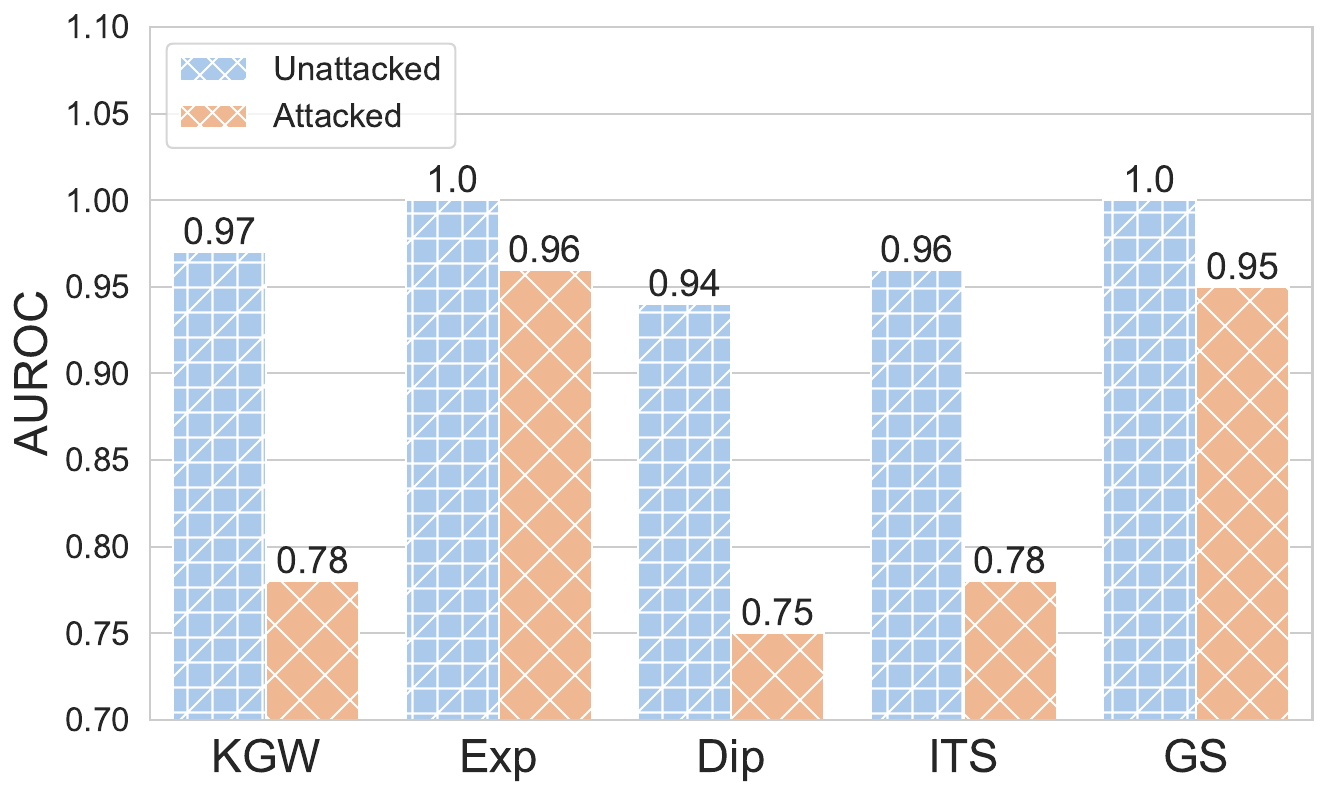}
\caption{Comparison of the robustness of decoding-based watermark on Completion task. Blue histograms indicate unattacked conditions and red histograms show attacked scenarios. The AUROC is calculated for 40 detection tokens, with GumbelSoft set at a 0.3 temperature. \textbf{Exp}, \textbf{Dip}, and \textbf{GS} refer to Exponential, Dipmark, and GumbelSoft, respectively. GumbelSoft and Exponential show higher robustness when facing the T5-span attack.}
\label{fig:robustness_com}
\end{figure}

GumbelSoft watermark exhibits the highest detectability, likely explained by the expectation and variation theory in Theorem~\ref{theorem:expectation of GumbelMax watermark}. Increased detection token amounts also improve detectability, aligning with findings from \citet{Chakraborty_Bedi_Zhu_An_Manocha_Huang_2023}. The high-entropy Llama2-7b model in Completion tasks shows greater detectability than the lower entropy Llama2-7b-chat in QA tasks, as high entropy facilitates easier watermark embedding.
Regarding generation quality (perplexity), GumbelSoft shows relatively low perplexity. In contrast, the KGW watermark's biased logits modification leads to high perplexity, while Dipmark's strategy of amplifying high-probability tokens results in the lowest perplexity in the Completion task. For the QA task, the low perplexity across all methods, attributed to the low entropy of Llama2-7b-chat, diminishes the value of comparative perplexity analysis.
\subsection{Robustness}
\begin{figure}[ht]
\centering
\includegraphics[width=1.0\linewidth]{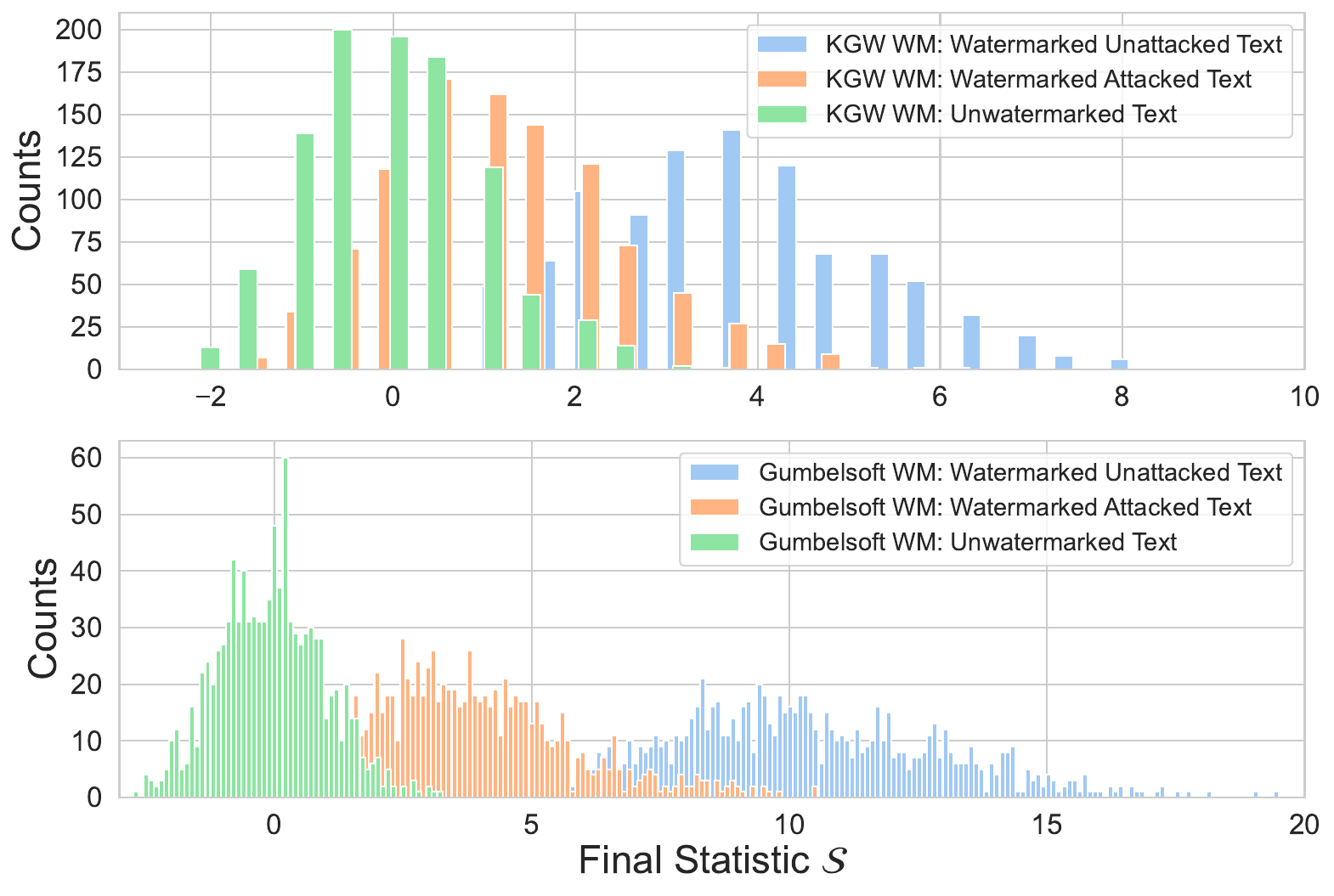}
\caption{Comparison of final statistic for KGW and GumbelSoft watermark on Completion task. The final statistic is calculated for 40 detection tokens, with GumbelSoft set at the temperature of 0.3. The robustness of GumbelSoft stems from the strong pattern of the GM watermark, ensuring a large gap in the Final statistic between watermarked text and natural text.
}
\label{fig:robustness_final_statistic}
\end{figure}
In this section, we assess the robustness of various decoding-based watermarking schemes, with results for the Completion task in Figures~\ref{fig:robustness_com} and for the QA task in Appendix~\ref{Appendix:robustness}. All texts, both watermarked and unwatermarked, were tested under the T5-span attack (explained in Appendix~\ref{Appendix:robustness}).

Our key finding reveals that the Exponential and GumbelSoft watermarks are particularly robust against the T5-span attack, in contrast to other watermarks. Their AUROC values, as well as FPR and FNR metrics, remained stable post-attack, while other schemes experienced significant declines. This robustness can be attributed to the effective embedding of watermarks by the GumbelMax-trick, ensuring significant final statistics despite per-token score alterations. Further analysis, involving a comparative study of final statistic distributions between KGW and GumbelSoft watermarks, is shown in Figure~\ref{fig:robustness_final_statistic}. The results demonstrate that while attacked watermarked texts under KGW show considerable overlap with unwatermarked texts, our GumbelSoft watermark displays less overlap, indicating its greater robustness.

\section{Conclusion}
\label{section:conclusion}
We observed that the GumbelMax-trick-based watermark(GM watermark) produces identical responses to identical queries due to the deterministic nature of both the Decoder and Pseudo-random functions. To address this, we introduce three diversified variants aimed at enhancing GM watermark diversity. Furthermore, we question the need for an Exponential transformation~\citep{Aaronson_Kirchner} in watermark embedding and propose a new approach named Logits-Addition watermark.
Our experiments across these variants for both Exponential and Logits-Addition watermarks identified GumbelSoft, a softmax-based Logits-Addition variant, as the optimal choice. Comparative analysis with other decoding-based watermarks demonstrated that GumbelSoft surpasses in detectability, maintains lower perplexity, and ensures higher robustness.

\section*{Limitations}
\label{section:Limitations}
GumbelSoft watermark's Ngram pseudo-random function is susceptible to paraphrase attacks due to its dependence on the previous h tokens for key determination. In terms of quality assessment, we solely rely on perplexity, whereas some studies utilize downstream tasks for evaluation. Our mathematical analysis is focused solely on the Logits-Addition watermarking technique, we do not provide a comprehensive mathematical analysis of the GumbelSoft watermark.

\section*{Ethical Considerations}
\label{section:ethical}
As advanced language models increasingly demonstrate remarkable capabilities, concerns regarding their misuse have escalated. Consequently, the development of effective methods for detecting machine-generated text has become crucial. The GM watermark has emerged as a highly effective technique for differentiating between machine-generated and natural text. Nevertheless, the GM watermark is limited by issues of diversity, which may hinder its practical application. The GumbelSoft watermark represents a straightforward yet effective strategy to address this limitation. This approach maintains the watermark's detectability while significantly enhancing its generative diversity. We believe that our method will facilitate the broader implementation of the GM watermark in various applications.

\section*{Acknowledgement}
\label{section:ack}
We would like to express our gratitude to our colleagues at Fudan University, especially to Yikai Zhang, Caiyu Hu, and Wei Shi for their valuable comments and spirited discussions on this project. This research is funded by the Science and Technology Commission of Shanghai Municipality Grant (No. 22511105902)

\bibliography{custom}
\newpage
\newpage
\appendix
\section{Baselines}
\label{proof:baselines}
Here, we present a consolidated mathematical representation within a unified taxonomy for the baseline watermark schemes.
\subsection{KGW Scheme}
For the KGW scheme, as proposed by \citet{pmlr-v202-kirchenbauer23a}:
\begin{itemize}
\item \textbf{Context}: The previous $h$ tokens.
\item \textbf{Pseudo-random Function}: $F_{sk}(\text{context})$ hashes the context to seed, then uses this seed to generate a random vector in $\{0,1\}^{|\mathcal{V}|}$, the vector has $\gamma|\mathcal{V}|$ 1's and $(1-\gamma)|\mathcal{V}|$ 0's.
\item \textbf{Decoder}: $\Gamma(\xi_t,l_t)$ samples a token from $\textrm{softmax}(\delta*\xi_t+l_t)$.
\item \textbf{Scorer}: $\phi(\xi_t,w_t)=\xi_t[w_t]$.
\item \textbf{Statistic Aggregator}: $$\Phi(s_1,\ldots,s_T)=\frac{\sum_{t=1}^T s_t-\gamma T}{\sqrt{T\gamma(1-\gamma)}}$$
\end{itemize}
\subsection{Exponential Scheme}
For the Exponential scheme, as proposed by \citet{Aaronson_Kirchner}:
\begin{itemize}
\item \textbf{Context}: The previous $h$ tokens.
\item \textbf{Pseudo-random Function}: $F_{sk}(\text{context})$ hashes the context to a seed, then uses this seed to generate a random vector in $(0,1)^{|\mathcal{V}|}$, each element is uniformly sample from (0,1).
\item \textbf{Decoder}: $\Gamma(\xi_t,l_t)=\argmax_{1\leq i\leq |\mathcal{V}|}\frac{\log\xi_{t}[i]}{p_{t}[i]}$, where $p_t=\textrm{softmax}(l_t)$.
\item \textbf{Scorer}: $\phi(\xi_t,w_t)=-\log(1-\xi_t[w_t])$.
\item \textbf{Statistic Aggregator}: $$\Phi(s_1,\ldots,s_T)=\frac{1}{\sqrt{T}}\sum_{t=1}^Ts_t - \sqrt{T}$$
\end{itemize}
\subsection{Dipmark Scheme}
For the Dipmark scheme, as proposed by \citet{wu2023dipmark}:
\begin{itemize}
\item \textbf{Context}: The previous $h$ tokens.
\item \textbf{Pseudo-random Function}: $F_{sk}(\text{context})$ hashes the context to a seed, then uses this seed to generate a random permutation on the vocabulary $\mathcal{V}$.
\item \textbf{Decoder}: $\Gamma(\xi_t,l_t)$ samples token $\xi_t[i]$ with probability $\lambda(i)-\lambda(i-1)$, where $\lambda(i)=\max\{\sum_{j=1}^i p_t(\xi_t[j])-\alpha,0\}+\max\{\sum_{j=1}^i p_t(\xi_t[j])-(1-\alpha),0\}$, where $p_t$=softmax($l_t$).
\item \textbf{Scorer}: $\phi(\xi_t,w_t)=\mathbf{1}_{\{w_t\in\xi_t[\gamma |\mathcal{V}|:|\mathcal{V}|]\}}$.
\item \textbf{Statistic Aggregator}: $$\Phi(s_1,s_2,\ldots,s_T)=\frac{\sum_{t=1}^T s_t-(1-\gamma) T}{\sqrt{T}}$$
\end{itemize}
\subsection{ITS Scheme}
For the ITS scheme, as proposed by \citet{kuditipudi2023robust}:
\begin{itemize}
\item \textbf{Context}: A global watermark key sequence $\xi$-list and the position index $t$. Each watermark key $\xi$-list[i] consists of a permutation $\pi$ on the vocabulary and a random number $\mu$ in $(0,1)$.
\item \textbf{Pseudo-random Function}: $F_{sk}(\text{context})=\xi$-list[t].
\item \textbf{Decoder}: $\Gamma(\xi_t,l_t)=\pi^{-1}(\min\{\pi(i):p_t(\{j:\pi(j)\leq\pi(i)\})\geq\mu\})$, where $\xi_t=(\mu,\pi)$ and $p_t=\textrm{softmax}(l_t)$.
\item \textbf{Scorer}: $\phi(\xi_t,w_t)=|\mu-\eta(\pi(w_t))|$, where $\eta(k)=\frac{k-1}{|\mathcal{V}|-1}$.
\item \textbf{Statistic Aggregator}: $$\Phi(s_1,s_2,\ldots,s_T)=-\frac{1}{T}\sum_{t=1}^T s_t$$
\end{itemize}
\subsection{Design Principles}
We now explore the design principles underlying these fundamental components.
\paragraph{Context}
For Watermark generators and detectors, it is essential to recognize that they share the same context, which is constrained to the previous tokens of $w_t$. This limitation arises from the auto-regressive nature of the Watermark generator, which sequentially generates tokens from left to right. A conventional approach for context selection is to use the previous h tokens: $w_{t-h}, \ldots, w_{t-1}$. However, this design is vulnerable to paraphrase attacks, as such attacks can alter these preceding tokens, subsequently modifying the watermark key $\xi_t$, and ultimately affecting the per-token score $s_t$. A more robust approach involves considering the semantic meaning of previous tokens, based on the rationale that paraphrasing maintains the semantics despite changing the tokens~\cite{liu2023semantic}. \citet{kuditipudi2023robust} suggest utilizing a global list for storing all watermark keys and retrieving a specific watermark key using the position index $t$
\paragraph{Pseudo-random Function.}
The pseudo-random function's role is to determine the watermark key $\xi_t$
based on the given context. This function could be as basic as a hash function of the context or might involve leveraging an embedding model to extract the context's semantic content. An alternative method is to use the position index $t$ to retrieve a watermark key from a global list. It is crucial to note that both the Watermark generator and detector share the same pseudo-random function.
\paragraph{Decoder.}
The decoder is integral to the Watermark generator, utilizing the watermark key $\xi_t$ and the logits vector $l_t$ to determine the subsequent token $w_t$. Implementation methods for this component vary among different watermarks.
\paragraph{Scorer.}
The scorer is to establish a correlation between the watermark key $\xi_t$ and the token $w_t$. Nevertheless, using a global watermark key list and the position index $t$ for key retrieval can result in a significant alignment shift issue. This problem manifests as a misalignment between the watermark key $\xi_t$ and the token $w_t$ in texts subjected to insertion or deletion attacks. To address this, \citet{kuditipudi2023robust} recommend using alignment cost or edit distance for computing the sequence score, as opposed to the per-token score.
\paragraph{Statistic Aggregator.}
Finally, the statistic aggregator compiles all per-token scores or employs a single sequence score to ascertain the presence of a watermark. A typical method involves calculating the z-score and p-value of collected scores. Alternatively, one could use the empirical cumulative distribution function~\cite{kuditipudi2023robust} for final statistical analysis.

\section{Mathematical Proofs}
\subsection{Unbiasedness for GumbelMax}
We demonstrate that the GumbelMax-trick is mathematically equivalent to directly sampling from the categorical distribution $\pi$, thereby establishing its unbiased nature when utilized in text watermarking applications. 

Denote the vocabulary as $\mathcal{V}$, the vector of logits as $l=(l_1,\ldots,l_{|\mathcal{V}|})$, and a sequence of independent Gumbel-distributed random variables as $\xi_1,\ldots,\xi_{|\mathcal{V}|} \sim\textrm{Gumbel}(0,1)$. 
\label{proof:Gumbel_unbiased}
\begin{align}
    & \mathbb{P}_{\xi\sim \textrm{Gumbel}(0,1)^{|\mathcal{V}|}}\left[\argmax_{1\leq i\leq|\mathcal{V}|}\{\xi_i+l_i\}=x\right] \\ 
    &= \mathbb{P}_{\eta\sim Q(\cdot)}\left[\argmax_{1\leq i\leq|\mathcal{V}|}\eta_i=x\right] \\ 
    &= \mathbb{P}_{\eta\sim Q(\cdot)}\left[\eta_x \geq \eta_i,\forall i\neq x\right] \\
    &=\mathop{\mathbb{E}}\limits_{Y\sim q(\cdot)}\left[\prod_{i\neq x} \mathbb{P}\left[Y\geq\eta_i\right]\right] \\
    &= \int_{-\infty}^{+\infty} f(y-l_x) \prod_{i\neq x} F(y-l_i)dy \\
    &= \int_{-\infty}^{+\infty} e^{-((y-l_x)+e^{-(y-l_x)})}\prod_{i\neq x}e^{-e^{-(y-l_i)}}dy \\
    &=\int_{-\infty}^{+\infty}e^{\sum_{i=1}^{|\mathcal{V}|}-e^{l_i-y}}e^{l_x-y}dy \\
    &=\int_{-\infty}^{+\infty}e^{-e^{-y}\sum_i e^{l_i}}e^{-y}e^{l_x}dy \\
    &=Zp_x\int_{-\infty}^{+\infty}e^{-Ze^{-y}}e^{-y}dy \\
    &= Zp_x\frac{1}{Z} \\
    &= p_x
\end{align}
In equation (2), we introduce a variable substitution $\eta_i=\xi_i+l_i$ for simplification. Moving to equation (4), we define the random variable $Y$ to represent $\eta_x$ for enhanced clarity, and we assume that $Y$ follows the distribution $q(.)$. Furthermore, we utilize the independence of $\eta_i, i=1,\ldots,|\mathcal{V}|$. In equation (5), $f(.)$ denotes the probability density function of the Gumbel(0,1) distribution, while $F(.)$ represents its cumulative distribution function. Finally, in equation (9), we introduce $Z=\sum_i e^{l_i}$ as a notation simplification.

\subsection{Equivalence of Two Representations}
We contend that the token generation processes for the Logits-Addition watermark and the Exponential watermark are mathematically equivalent, though their respective per-token scoring mechanisms differ. To illustrate this equivalence, we present the following equations, equation (12) defines the Logits-Addition watermark, while equation (15) corresponds to the definition of the Exponential watermark.
\label{proof:Gumbel_equivalent}
\begin{align}
  w &=\argmax_{1\leq i\leq|\mathcal{V}|}\{\eta_i+l_i\}\\
    &=\argmax_{1\leq i\leq|\mathcal{V}|} e^{\eta_i+l_i}\\
    &=\argmax_{1\leq i\leq|\mathcal{V}|}\frac{-p_i}{\log\xi_i}\\
    &=\argmax_{1\leq i\leq|\mathcal{V}|}\frac{\log\xi_i}{p_i}
\end{align}
Here, we utilize the relationship $p_i=\textrm{softmax}(l_i)\propto e^{l_i}$ and $\eta_i=-\log(-\log\xi_i)$. In these notations, we omit the position index t for simplicity, and we assume $\eta_i\sim \textrm{Gumbel}(0,1)$ and $\xi_i\sim\textrm{Uniform}(0,1).$

While the token generation processes for the Logits-Addition watermark and the Exponential watermark are equivalent, their scoring methods are distinct:
\begin{align}
  w &= \eta[w] \\
    &= -\log(-\log\xi[w]) \\
    &\neq -\log(1-\xi[w])
\end{align}
Equation (16) specifies the per-token scoring for the Logits-Addition watermark while equation (18) is the scoring method for the Exponential watermark.

\subsection{Expectation and Variance for Per-token Score}
\label{proof:Gumbel_expectation}
We now establish the expected per-token score for texts, distinguishing between those with and without the Logits-Addition watermark. In the case of unwatermarked texts, the watermark token, $w_t$, exhibits no correlation with $\xi_t$. Consequently, $\xi_t[w_t]$ adheres to a Gumbel(0,1) distribution. This leads to
\begin{equation*}
    \begin{aligned}
    \mathbb{E}[s_t]&=\mathbb{E}[\xi_t[w_t]]=\gamma,\\
    \textrm{Var}[s_t]&=\textrm{Var}[\xi_t[w_t]]=\frac{\pi^2}{6}.
    \end{aligned}
\end{equation*}
Conversely, for watermarked texts, a correlation exists between $w_t$ and $\xi_t$. This correlation alters the distribution of $\xi_t[w_t]$, diverging it from the standard Gumbel(0,1) form. To compute its expected value, we define $\xi_t[w_t]$ as a random variable X. We then deduce its cumulative distribution function (CDF), $F(x)$, and probability density function (PDF), $f(x)$. Utilizing this PDF, we calculate the expectation of X. For simplification, we exclude the position index `t' in the subsequent equations. Here, $\xi_i$ represents $\xi[i]$, implying $\xi_w$ is equivalent to $\xi_t[w_t]$, and similar conventions apply to other notations.
\begin{align}
  F(x)&= \mathbb{P}\left[X\leq x\right] \\
      &= \mathbb{P}\left[\xi_w\leq x\right] \\
      &= \mathbb{P}\left[\xi_i+l_i-l_w\leq x,\forall i\right] \\
      &= \mathbb{P}\left[e^{\xi_i+l_i-l_w}\leq e^x,\forall i\right]\\
      &= \mathbb{P}\left[\frac{-1}{\log h_i}\frac{p_i}{p_w}\leq e^x,\forall i\right]\\
      &= \mathbb{P}\left[\frac{p_i}{p_w}e^{-x}\leq-\log h_i,\forall i\right]\\
      &= \prod_{i=1}^{|\mathcal{V}|} \mathbb{P}\left[\frac{p_i}{p_w}e^{-x}\leq -\log h_i\right]\\
      &= \prod_{i=1}^{|\mathcal{V}|} 1-\mathbb{P}\left[-\log h_i\leq \frac{p_i}{p_w}e^{-x}\right]\\
      &= \prod_{i=1}^{|\mathcal{V}|} 1-(1-e^{-\frac{p_i}{p_w}e^{-x}})\\
      &= \prod_{i=1}^{|\mathcal{V}|} e^{-\frac{p_i}{p_w}e^{-x}}\\
      &= e^{\sum_{i=1}^{|\mathcal{V}|}\frac{-p_i}{p_w}e^{-x}} \\
      &= e^{\frac{-e^{-x}}{p_w}}
\end{align}
In equation (22), we utilize the fact that $p_i\propto e^{l_i}$ and $\xi_i=-\log(-\log h_i)$. Equation (24) leverages the independence of $h_i$. Equation (26) uses the fact that $-log h_i\sim \textrm{Exp}(1)$. Finally, equation (29) employs the fact that $\sum_{i=1}^{|\mathcal{V}|} p_i = 1$.
Hence, the density function:
$$f(x)=F^{'}(x)=\frac{e^{-x}}{p_w} e^{\frac{-e^{-x}}{p_w}}$$
The expectation:
\begin{align}
\mathbb{E} [\xi_t[w_t]] &= \mathbb{E} [X]\\
                  &=\int_{-\infty}^{+\infty} xf(x) dx\\
                  &=-\int_{-\infty}^{+\infty} x\frac{e^x}{p_w} e^{\frac{-e^x}{p_w}} dx \\
                  &=-\frac{1}{p_w}[p_w\log p_w-\gamma p_w]\\
                  &=-\log p_w+\gamma
\end{align}
The equation (32) use the fact that:
$$\int_{-\infty}^{+\infty} x e^x e^{\frac{-e^x}{t}}dx=t\log t-\gamma t$$
We now prove the fact. This is not a standard integral that can be solved by elementary functions. However, we can attempt to solve it using the substitution method and some properties of the Gamma and incomplete Gamma functions, which are commonly used to handle integrals involving exponentials of exponentials.
\begin{align}
& \int_{-\infty}^{+\infty} x e^x e^{\frac{-e^x}{t}}dx \\
&= \int_{0}^{+\infty} \log(u)e^{-u/t}du \\
&= \int_{0}^{+\infty} \log(vt)e^{-v}t dv \\
&= t\int_{0}^{+\infty} (\log(v)+\log(t))e^{-v} dv \\
&= t\log(t)\int_0^{+\infty}e^{-v} dv + t\int_0^{+\infty}\log(v)e^{-v}dv \\
&= t\log(t)+t\int_0^{+\infty}\log(v)e^{-v}dv\\
&= t\log(t)-\gamma t
\end{align}
In Equation(35), we use variable substitution $u=e^x$, In Equation(36), we use variable substitution $v=\frac{u}{t}$, In Equation (39), we use the definition of Euler-Mascheroni constant:
$\gamma=-\int_0^{+\infty}\log(v)e^{-v}dv$.

As for the variance of the per-token score for watermarked text, we can also derive it via the probability density function $f(x)$.
\begin{align}
&\textrm{Var}[s_t]\\
&=\textrm{Var}[\xi_t[w_t]]\\
&=\mathbb{E}[X^2]-(\mathbb{E}[X])^2\\
&=\int_{-\infty}^{+\infty} x^2f(x)dx-(-\log p_w+\gamma)^2\\
&=\int_{-\infty}^{+\infty} x^2\frac{e^{-x}}{p_w} e^{\frac{-e^{-x}}{p_w}}dx-(-\log p_w+\gamma)^2\\
&\leq \frac{2{p_w}^2}{(1-p_w)^3}+\frac{2}{p_w}-(-\log p_w+\gamma)^2
\end{align}

In equation(48), we use the fact that $$\int_{-\infty}^{+\infty} x^2\frac{e^{-x}}{p_w} e^{\frac{-e^{-x}}{p_w}}dx\leq\frac{2{p_w}^2}{(1-p_w)^3}+\frac{2}{p_w}$$
We now prove it:
\begin{align}
&\int_{0}^{+\infty}  x^2\frac{e^{-x}}{p_w} e^{\frac{-e^{-x}}{p_w}}dx\\
&\leq \int_{0}^{+\infty}  x^2\frac{e^{-x}}{p_w} dx\\
&=\frac{2}{p_w}
\end{align}

\begin{align}
&\int_{-\infty}^0  x^2\frac{e^{-x}}{p_w} e^{\frac{-e^{-x}}{p_w}}dx\\
&=\int_0^{+\infty}  x^2\frac{e^{x}}{p_w} e^{\frac{-e^{x}}{p_w}}dx\\
&=\int_0^{+\infty}  \frac{x^2}{p_w} e^{(x-\frac{e^{x}}{p_w})}dx\\
&\leq\int_0^{+\infty}  \frac{x^2}{p_w} e^{(1-\frac{1}{p_w})x}dx\\
&=\frac{2{p_w}^2}{(1-p_w)^3} 
\end{align}

A similar theorem also holds for the Exponential watermark.
For unwatermarked text:
$$\mathbb{E}[s_t]=\mathbb{E}[-\log(1-\xi_t[w_t])]=1$$
$$\textrm{Var}[s_t]=\textrm{Var}[-\log(1-\xi_t[w_t])]=1$$
For watermarked text,
\begin{equation*}
    \begin{aligned}
    \mathbb{E}[s_t]&=\mathbb{E}[-\log(1-\xi_t[w_t])]\\
    &\geq 1+(\frac{\pi^2}{6}-1)(-p_w \log p_w),\\
    \textrm{Var}[s_t]&=\textrm{Var}[-\log(1-\xi_t[w_t])]\\
    &=\psi_1(1)-\psi_1(1+\frac{1}{p_w}),
    \end{aligned}
\end{equation*}
where $\psi_1$ is the trigamma function. The proof can be found in ~\citet{fernandez2023bricks}

\begin{table*}[t]
\centering
\small
\begin{tabular}{ccsssdddq}
\toprule
& & \multicolumn{3}{c}{\textbf{Diversity}}  & \multicolumn{3}{c}{\textbf{Detectability}} & \multicolumn{1}{c}{\textbf{Quality}} \\
\cmidrule(lr){3-5} \cmidrule(lr){6-8} \cmidrule(lr){9-9}
& & \multicolumn{1}{c}{\textbf{Self-Bleu} $\downarrow$}  & \multicolumn{1}{c}{\cellcolor{white}\textbf{Dist-1} $\uparrow$}&   \multicolumn{1}{c}{\textbf{Dist-2} $\uparrow$} &  \multicolumn{1}{c}{\textbf{AUROC} $\uparrow$} &  \multicolumn{1}{c}{\textbf{FPR} $\downarrow$ }&  \multicolumn{1}{c}{\textbf{FNR} $\downarrow$ }&  \multicolumn{1}{c}{\textbf{PPL} $\downarrow$ } \\

\midrule
\multirow{15}{*}{\rotatebox{90}{\textbf{Exponential}}}
&vanilla & 1.000 & 0.010 & 0.014 & 1.000 & 0.000 & \textbf{0.000} & \textbf{10.953} \\
\cdashlinelr{2-9}
&drop\_prob=0.05 & 0.367 & 0.222 & 0.529 & 1.000 & 0.000 & \textbf{0.000} & 11.450 \\
&drop\_prob=0.10 & 0.227 & 0.254 & 0.633 & 1.000 & 0.000 & 0.001 & 11.423 \\
&drop\_prob=0.20 & 0.146 & 0.300 & 0.733 & 1.000 & 0.000 & 0.001 & 11.839 \\
&drop\_prob=0.30 & 0.113 & 0.307 & 0.766 & 1.000 & 0.000 & 0.002 & 11.911 \\
&drop\_prob=0.40 & \textbf{0.087} & 0.317 & 0.788 & 1.000 & 0.001 & 0.005 & 11.964 \\
\cdashlinelr{2-9}
&shift\_max=10 & 0.991 & 0.079 & 0.146 & 0.999 & 0.000 & 0.003 & 11.307 \\
&shift\_max=30 & 0.798 & 0.158 & 0.333 & 0.999 & 0.000 & 0.002 & 11.084 \\
&shift\_max=50 & 0.645 & 0.184 & 0.403 & 1.000 & 0.000 & 0.003 & 11.222 \\
&shift\_max=100 & 0.414 & 0.221 & 0.496 & 0.999 & 0.000 & 0.003 & 11.102 \\
&shift\_max=200 & 0.247 & 0.235 & 0.546 & 0.999 & 0.000 & 0.004 & 11.068 \\
\cdashlinelr{2-9}
&soft\_temp=0.1 & 0.388 & 0.210 & 0.490 & 1.000 & 0.000 & \textbf{0.000} & 11.218 \\
&soft\_temp=0.2 & 0.244 & 0.238 & 0.565 & 1.000 & 0.000 & 0.001 & 11.353 \\
&soft\_temp=0.3 & 0.202 & 0.265 & 0.630 & 1.000 & 0.000 & 0.001 & 11.610 \\
&soft\_temp=0.4 & 0.169 & 0.275 & 0.669 & 1.000 & 0.000 & 0.001 & 11.874 \\
&soft\_temp=0.5 & 0.146 & 0.285 & 0.686 & 1.000 & 0.000 & 0.001 & 12.222 \\
\midrule
\multirow{15}{*}{\rotatebox{90}{\textbf{Logits-Addition}}}
&vanilla & 1.000 & 0.010 & 0.014 & 1.000 & 0.000 & 0.000 & 10.953  \\
\cdashlinelr{2-9}
&drop\_prob=0.05 & 0.421 & 0.205 & 0.493 & 0.999 & 0.000 & 0.003 & 11.561 \\
&drop\_prob=0.10 & 0.209 & 0.268 & 0.652 & 0.999 & 0.000 & 0.003 & 11.754 \\
&drop\_prob=0.20 & 0.143 & 0.292 & 0.729 & 0.999 & 0.000 & 0.005 & 11.997 \\
&drop\_prob=0.30 & 0.097 & 0.309 & 0.774 & 0.999 & 0.001 & 0.005 & 11.890 \\
&drop\_prob=0.40 & 0.093 & \textbf{0.319} & \textbf{0.790} & 0.999 & 0.003 & 0.006 & 12.156 \\
\cdashlinelr{2-9}
&shift\_max=10 & 0.991 & 0.078 & 0.144 & 0.999 & 0.000 & 0.006 & 11.228 \\
&shift\_max=30 & 0.806 & 0.159 & 0.335 & 0.998 & 0.002 & 0.006 & 11.243 \\
&shift\_max=50 & 0.627 & 0.188 & 0.412 & 0.998 & 0.000 & 0.006 & 11.250 \\
&shift\_max=100 & 0.417 & 0.220 & 0.497 & 0.998 & 0.000 & 0.006 & 11.536 \\
&shift\_max=200 & 0.246 & 0.242 & 0.559 & 0.998 & 0.001 & 0.007 & 11.243 \\
\cdashlinelr{2-9}
&soft\_temp=0.1 & 0.370 & 0.213 & 0.497 & 1.000 & 0.000 & 0.001 & 11.159 \\
&soft\_temp=0.2 & 0.227 & 0.243 & 0.581 & 1.000 & 0.000 & 0.002 & 11.309 \\
&soft\_temp=0.3 & 0.158 & 0.254 & 0.608 & 1.000 & 0.000 & 0.001 & 11.820 \\
&soft\_temp=0.4 & 0.121 & 0.276 & 0.661 & 1.000 & 0.000 & 0.001 & 12.831 \\
&soft\_temp=0.5 & 0.100 & 0.298 & 0.699 & 1.000 & 0.000 & 0.001 & 14.140 \\
\bottomrule
\end{tabular}
\caption{Comparison of three variants of both Exponential and Logits-Addition watermarks in the Completion task. The variants include \textbf{drop\_prob=0.2}, sampling from the language model directly at a 0.2 probability; \textbf{shift\_max=100}, where the watermark key is cyclically shifted within a 0-100 range; and \textbf{soft\_temp=0.3}, which uses a softmax sampling with a temperature of 0.3 to balance randomness. \textbf{Vanilla} is the original two GumbelMax watermarks without any technique to enhance diversity. The detectability is measured by 100 detection tokens. Note that Logits-Addition+soft\_temp is our GumbelSoft watermark.}
\label{table:diversity_com}
\end{table*}

\section{Experiment details}
We describe all experiment details here. We run all experiments five times and report the average value. The variance of the five repeated experiments is negligible, approaching zero; therefore, it has been excluded from the table. The detailed comparison of
our GumbelSoft watermark with other watermark variants in the Text Completion task is presented in Table \ref{table:diversity_com}.

\subsection{Diversity}
\label{Appendix:diversity}
We began by carefully selecting 40 high-entropy prompts to elicit a wide range of completions. These prompts were split evenly into two categories: 20 prompts followed a Completion format tailored for Llama2-7b, while the remaining 20 were structured in a QA format, specifically designed for Llama2-7b-chat. Each prompt was queried 50 times, and we assessed the resulting completions using metrics such as Self-Bleu, Distinct 1-gram, and Distinct 2-gram. The average values of these metrics were then computed for the 20 prompts in each category. We control the max generation length for each prompt to be 256 tokens.

For the soft\_temp parameter, we tested five different temperature settings: 0.1, 0.2, 0.3, 0.4, and 0.5. In the case of the shifted watermark key, we experimented with five maximum shift values: 10, 30, 50, 100, and 200. For drop probability, the tested probabilities were 5\%, 10\%, 15\%, 20\%, and 40\%. We evaluated detectability and quality using a sample of 100 generated tokens, while diversity assessments were conducted with a sample size of 256 tokens.
\subsection{Detectability}
\label{Appendix:detectability}
The objective of text watermarking is to embed a concealed pattern into generated texts and subsequently detect this pattern to ascertain if the text is watermarked. We gathered 1,000 lengthy texts from the news-like validation subset of the C4 dataset, dividing each text into two parts: the first 50 words as prompt and the remaining as gold-completion. For each watermarking scheme, we utilized Llama2-7b to create both watermarked and unwatermarked completions for these 1,000 prompts. The effectiveness of each scheme was then assessed using the corresponding detector to evaluate 2,000 completions. Key metrics reported include AUROC (Area Under the Receiver Operating Characteristic), FPR (False Positive Rate) at a fixed FNR (False Negative Rate) of 0.01, and FNR at a fixed FPR of 0.01. 

Additionally, we compiled 1,000 lengthy texts from the alpaca dataset. Unlike the C4 dataset, here we used only the question as a prompt to query Llama2-7b-chat, with the subsequent detection process mirroring that of the C4 dataset.

In line with the detectability theorem by \citet{Chakraborty_Bedi_Zhu_An_Manocha_Huang_2023}, we anticipate higher detectability in longer texts. Therefore, we report detection metrics for generated token lengths of 40, 60, 80, and 100. However, for quality assessment, we calculate perplexity only for texts with 100 generated tokens, as fewer tokens would inadequately represent quality measures. We use llama2-13b and llama2-13b-chat to evaluate ppl for the texts generated by llama2-7b and llama2-7b-chat respectively.

The hyper-parameters employed for each watermarking scheme are specified as follows: All experiments are conducted at a temperature setting of 1, except the GumbelSoft, which utilized a temperature setting of 0.3 to achieve an equilibrium between detectability and generation diversity. For KGW, we adopt $\delta=2$ and $\gamma=0.1$, following the recommendations by \citet{pmlr-v202-kirchenbauer23a}. For Dipmark, the parameters are set to $\alpha=0.45$ and $\gamma=0.5$, by \citet{wu2023dipmark}. Regarding ITS, we utilize a sample of 500 texts from the C4 subset for the computation of reference scores. We repeat the experiment 5 times to calculate the average value for each metric.

\begin{table*}[ht]
\centering
\resizebox{0.95\textwidth}{!}{ 
\small
\begin{tabular}{ccsssdddqqq}
\toprule
& & \multicolumn{3}{c}{\textbf{\# tokens=40}}  & \multicolumn{3}{c}{\textbf{\# tokens=60}} & \multicolumn{3}{c}{\textbf{\# tokens=100}} \\
\cmidrule(lr){3-5} \cmidrule(lr){6-8} \cmidrule(lr){9-11}& 
& \multicolumn{1}{c}{\textbf{AUROC} $\uparrow$}  
& \multicolumn{1}{c}{\cellcolor{white}\textbf{FPR} $\downarrow$} 
& \multicolumn{1}{c}{\textbf{FNR} $\downarrow$} 
& \multicolumn{1}{c}{\textbf{AUROC} $\uparrow$}  
& \multicolumn{1}{c}{\cellcolor{white}\textbf{FPR} $\downarrow$}
& \multicolumn{1}{c}{\textbf{FNR} $\downarrow$} 
& \multicolumn{1}{c}{\textbf{AUROC} $\uparrow$} 
& \multicolumn{1}{c}{\textbf{FPR} $\downarrow$ }
& \multicolumn{1}{c}{\textbf{FNR} $\downarrow$ }\\
 
\midrule
\multirow{6}{*}{\rotatebox{90}{\textbf{Completion}}}
& KGW & 0.580 & 0.989 & 0.981 & 0.598 & 0.990 & 0.971 & 0.632 & 0.979 & 0.957\\
& Exponential & \textbf{0.689} & 0.975 & \textbf{0.857} & 0.721 & 0.969 & \textbf{0.819} & 0.768 & 0.951 & \textbf{0.775}\\
& Dipmark & 0.553 & 0.984 & 0.986 & 0.570 & 0.985 & 0.983 & 0.589 & 0.985 & 0.976 \\
& ITS & 0.592 & 1.000 & 1.000 & 0.609 & 1.000 & 1.000 & 0.649 & 1.000 & 1.000 \\
& \textbf{GumbelSoft} & 0.685 & \textbf{0.971} & 0.862 & \textbf{0.723} & \textbf{0.966} & 0.841 & \textbf{0.770} & \textbf{0.951} & 0.781 \\
\bottomrule
\end{tabular}
}
\caption{A comparative analysis of the detectability across various decoding-based watermarking schemes under GPT-3.5 Turbo Paraphrase attack. The temperature for GumbelSoft is set to 0.3. All decoding-based watermarks perform poorly due to their reliance on previous contexts.}
\label{table:paraphrase_attack}
\vspace{1em}

\centering
\small
\begin{tabular}{lccc}
    \toprule
    \textbf{Methods} & \textbf{Average Cover Rate}$\uparrow$  & \textbf{Variance Cover Rate}$\downarrow$ & \textbf{Average PPL}$\downarrow$ \\
    \midrule
    Unwatermarked & 69.091 & 480.622 & 4.824 \\
    KGW & 70.015 & 368.197 & 5.695 \\
    Exponential & 70.396 & 264.503 & 4.919 \\
    Dipmark & 69.308 & 436.202 & 4.115 \\
    ITS & 70.137 & 382.446 & 4.855 \\
    \textbf{GumbelSoft} & 68.681 & 328.752 & 5.103 \\
    \bottomrule
\end{tabular}
\caption{Comparison of various watermarks on the constrained text generation task reveals that all watermarks perform similarly.}
\label{table:CTG}

\vspace{1em}
\begin{tabular}{lcccc}
\toprule
\textbf{Model} & \textbf{Average SBERT}$\uparrow$ & \textbf{Variance SBERT}$\downarrow$ & \textbf{Average Rouge1}$\uparrow$ & \textbf{Variance Rouge1}$\downarrow$ \\
\midrule
Unwatermarked & 0.619 & 0.036 & 0.219 & 0.006 \\
KGW & 0.637 & 0.029 & 0.225 & 0.006 \\
Exponential & 0.646 & 0.019 & 0.223 & 0.006 \\
Dipmark & 0.625 & 0.037 & 0.219 & 0.007 \\
ITS & 0.575 & 0.058 & 0.204 & 0.010 \\
\textbf{GumbelSoft} & 0.650 & 0.020 & 0.223 & 0.006 \\

\bottomrule
\end{tabular}
\caption{Comparison of various watermarks on the summarization task indicates that the GumbelSoft watermark performs slightly better than the others.}
\label{table:Summarize}
\end{table*}

\subsection{Robustness}
\label{Appendix:robustness}
\begin{figure}[ht]
\centering
\includegraphics[width=1.0\linewidth]{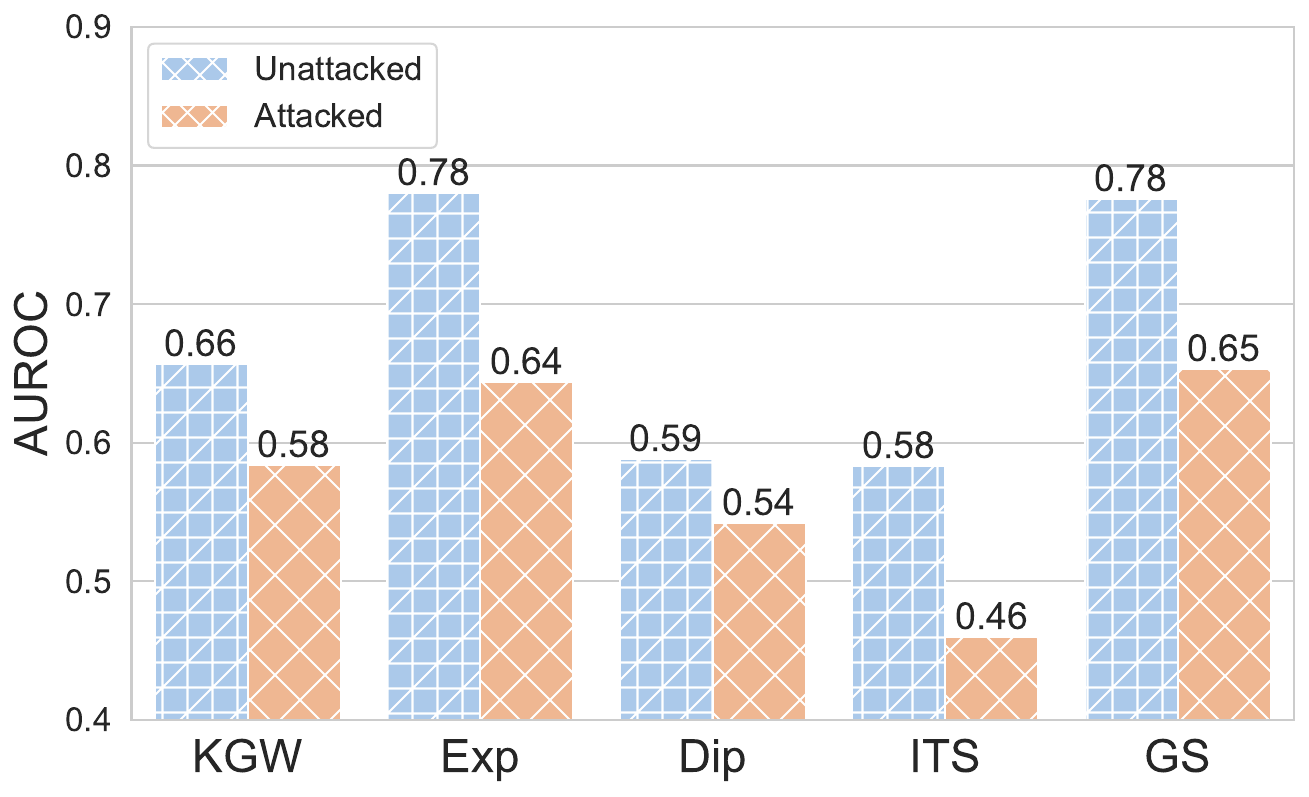}
\caption{Comparison of robustness of decoding-based watermark on QA task. Blue histograms indicate unattacked conditions and red histograms show attacked scenarios. The AUROC is calculated for 40 detection tokens, with GumbelSoft set at a 0.3 temperature. \textbf{Exp}, \textbf{Dip}, and \textbf{GS} refer to Exponential, Dipmark, and GumbelSoft, respectively.}
\label{fig:robustness_chat}
\end{figure}

\paragraph{T5-span Attack}
We employ the T5-span attack \cite{pmlr-v202-kirchenbauer23a} on both watermarked and unwatermarked texts. Each word in a text undergoes a potential attack with a probability of 0.5. For attacked words, we use their immediate five-word context (preceding and following) and apply t5-large \cite{Raffel_Shazeer_Roberts_Lee_Narang_Matena_Zhou_Li_Liu_2019} for context-based word prediction, replacing the original word with the predicted one. This process may occasionally retain the original word; however, we opt not to enforce unique substitutions to avoid excessive time consumption.

\paragraph{Paraphrase Attack}

We employ GPT-3.5 Turbo to perform paraphrase attacks on the completion task. The results, presented in Table~\ref{table:paraphrase_attack}, indicate that the paraphrase attack is significantly more potent than the T5-span attack due to its substantial modifications to the context. Consequently, all decoding-based watermarks exhibit poor performance under such an attack.

\subsection{Downstream Task Evaluation}
We further evaluate all decoding-based watermarks on downstream tasks to compare their generation quality.
\paragraph{Experiment Setting.}
We conducted quality assessment experiments on two sequence-to-sequence (seq2seq) downstream tasks: constrained text generation and summarization. In the constrained text generation task, the model receives a list of words and is tasked with generating a coherent sentence incorporating these words. In the summarization task, the model is given a long article (approximately 1500 tokens) and is expected to generate a concise summary. We utilized the Llama2-7b-chat model as our test model. For the constrained text generation and summarization tasks, we used the hard subset of CommonGen and CNN-DM datasets, respectively, each comprising 200 examples.

\paragraph{Metric and Prompt.}
For the constrained text generation task, we employed cover rate as the metric, which measures the percentage of provided words included in the generated sentence. The instruction prompt used was: ``Use the following words to create a sentence. The sentence should contain all provided words.'' For the summarization task, we utilized the F-measure of ROUGE-1 and cosine similarity of Sentence-BERT 'all-MiniLM-L6-v2' as metrics. The instruction prompt used was: "Please help me summarize the given article."

\paragraph{Results.}
The results are presented in Table~\ref{table:CTG} and Table~\ref{table:Summarize}. The data indicate that our Gumbelsoft model outperforms other decoding-based watermarks in terms of sentence-BERT cosine similarity for the summarization task. In the constrained text generation task, while all methods exhibit comparable performance, the extremely high variance diminishes the significance of the analysis.

\section{Responsible NLP Research.}
The C4 dataset is under the terms of ODC-BY and the Alpaca dataset is under the terms of Creative Commons NonCommercial (CC BY-NC 4.0). Our research fully obeys these licenses. C4 and Alpaca datasets are publicly available and do not contain private information for any individual.

\end{document}